\documentclass[journal]{IEEEtran}
\usepackage{graphicx}
\usepackage{amssymb,amsmath}
\usepackage{subfigure}
\usepackage{multirow}
\usepackage{algorithm}
\usepackage{algpseudocode}

\ifCLASSINFOpdf
\else
\fi

\usepackage{hyperref}

\begin{document}

%
\title{Spatial Stimuli  Gradient Sketch Model  }
%
%
%

\author{Joshin John Mathew, and Alex Pappachen James,\textit{ IEEE Senior Member}
\thanks{A.P. James  is a faculty with the Department
of Electrical and Electronics Engineering, Nazarbayev University,
Astana e-mail: (see \href{http://www.biomicrosystems.info/apj}{http://www.biomicrosystems.info/apj}).}
\thanks{J. J Mathew is a research staff member with Enview R\&D labs \textit{LLP}}
\thanks{Manuscript received December 1, 2014;revised January 26, 2015 }}

\maketitle

\begin{abstract}
The inability of automated edge detection methods inspired from primal sketch models to accurately calculate object edges  under the influence of pixel noise is an open problem.   Extending the principles  of image perception i.e. Weber-–Fechner  law, and Sheperd similarity law, we propose a new edge detection method and formulation that use  perceived brightness and neighbourhood similarity calculations in the determination of robust object edges. The robustness  of the detected edges  is benchmark  against Sobel, SIS, Kirsch, and Prewitt edge detection methods in an example face recognition problem showing statistically significant improvement in recognition accuracy and pixel noise tolerance.
\end{abstract}


\ifCLASSOPTIONpeerreview
\begin{center} \bfseries EDICS Category:
IMD-ANAL;
IMD-PATT \end{center}
\fi
%
\IEEEpeerreviewmaketitle

\section{Introduction}

\IEEEPARstart{T}{he} concept of spatial change detection is interrelated to the \textit{primal sketch theory} \cite{1,2,3} where the change in the intensities over a visual field is taken as feature than pixel intensities themselves. The primal sketch relates the geometry of the image intensity changes in the images to the early psychophysical mechanism of human eye. These edge models reflect to the general idea of spatial change detection \cite{14} from the neighbourhood of the pixel under consideration that find application as a feature preprocessing stage in general image bases pattern recognition problems \cite{15,16}. 

This paper presents a noise robust edge detection method using the local stimuli computed from the amount perceived brightness at each image location for  enhancing the quality of image inherent structural and intensity edge information. The proposed method computes a relative numerical quantification of the physical stimuli contributed to the image formation. Such computed local stimuli helps to extract and preserve   underlying structural edge details. The quality of the detected edges are demonstrated through examples of simulations with hypothetical edges and through standard face recognition problem.

\section{Background}
In digital images, object details are analyzed with the help of responses computed for intensity changes at its boundaries that are referred to as the edges \cite{6}. This primary idea is inspired by the biological vision processing of human eye and in its early form is formulated by primal sketch model \cite{2} describes an image using the image inherent structures based on the response computed with respect to edge formations using optimal smoothing filters and detection of intensity changes.

The primary criteria of any edge detection technique include its ability to provide best response to edges, good localization, continuity of edges and tolerance to image noise cum natural variance within the same regions. A comprehensive description of an image inevitably requires a robust retrieval of edge information by satisfying the criteria for edge detection. The quality of image description based on {primal sketch model} can be mathematically modeled with help of sketchable and unsketchable edge responses \cite{2}. The term sketchability is meant for useable edges and unsketchable edge responses are referred to the false edges formed due the intra-region variability together with the image noise. Primal structures retrieved comprising of all sketchable edges can be described as the ideal edge detection technique.
In real world images, it is not possible to perform an ideal detection, but to optimize for minimal trade-off between sketchable and unsketchable edges the best efforts are made by applying smoothing as the first stage of edge detection. Unfortunately, this  will contribute to reduction of many of the fine details due to the averaging operations used to implement the smoothing. This effect can be controlled up to certain degrees, by using specialized structure preserved denoising techniques such as non-local means filter \cite{4}, noise suppression filters \cite{17,18}, and   anisotropic diffusion filtering \cite{5}. On the other hand, hypothetically,   an edge response computation with inherent intra-region variability suppression can lead to more robust edge detection approaches. In this paper, the focus is  on the suppression of intensity variability other than edges in order to minimizes the unsketchable primitives.

\section{Proposed Method}
We propose to make use of Weber-Fechner' law \cite{7} to mathematically compute the amount of perceived brightness and the corresponding local stimuli. The aim is to enhance the inherent image features by weighting the original image using the magnitude of local stimuli resulting in the image formation. This magnitude of local stimuli is computed by finding the local variation in perceived brightness at the respective locations. Fechner law states that the amount of perceived brightness/loudness is equal to the logarithm the measured intensity of the physical stimulus. Thus, the perceived brightness, $B_p$ for a given image $I$, is expressed as

\begin{equation}
B_{p} = k\log_{10}(I)
\end{equation}

where $k$ is a constant.

The gradients of the perceived brightness would represent the edge information in the image. Since the gradient itself is a linear operator the noisy pixel
will not be suppressed by this operation. In addition, the gradient is a form of distance measure in context of measuring the localised inter-pixel variability. While there exists several of distance measures that are suitable to be used for detecting the spatial change \cite{14}, we propose to use Shepherd's similarity measure \cite{13} that would give a measure of perceived similarity on the measured distance. In addition to the  psychological dimension of the Shepherd's similarity, this measure  has the ability to suppress the outliers in the distance calculations due to the exponential nature of the transform.
Applying this theory to practise, localized spatial stimuli magnitude is numerically quantified as the net variation of perceived brightness along $x$ and $y$ directions.  The variation of $B$ along $x$ ($V_x$) and $y$ ($V_y$) directions is computed as a function of respective gradients ($g_x$ and $g_y$) and is expressed as:

\begin{equation}
V_x = g_x.\exp(-|g_x|) ~;~ 
V_y = g_y.\exp(-|g_y|)
\end{equation}

The magnitude of local stimuli ($V$) can be computed using Eq. 3.
\begin{equation}
V=\sqrt{V_{x}^2+V_{y}^2}
\end{equation}
The inherent image features by pixel wise weighting of original image (Fig. 1a) is computed using Eq. 3 to obtain the local stimuli magnitude map\footnote{It may be noted that the alternate form  $V_x=|g_x|(1-e^{-|g_x|})$ and $V_y=|g_y|(1-e^{-|g_y|})$ will be also able to amplify edges and suppress the noise, however, will have difficulty identifying slow varying and low contrast edges.} ($V$) (Fig. 1b).

\begin{figure}[hbtp]
\center
\subfigure[]{
  \includegraphics[width=.6in]{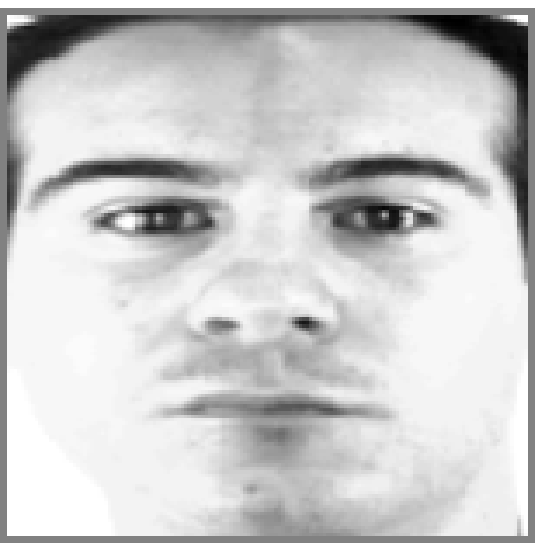}
   \label{fig1:subfig1}
   }
\subfigure[]{
  \includegraphics[width=.6in]{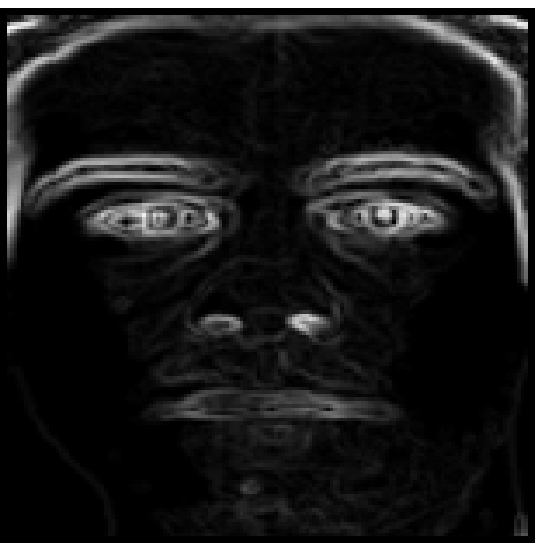}
   \label{fig1:subfig2}
   }
\caption{(a) Original image $I$, (b) Local stimuli magnitude map $V$ }
\label{figure1}
\end{figure}

\begin{algorithm}
        \caption{Local Stimuli Template}
        \label{localStimuliTemplate}
        \begin{algorithmic}[1]
                \Procedure{LST}{$I$}
                \State Read image, $I$.
                \State $B \gets k\log(I)$  // Perceived brightness\footnote{}  
                \State $[g_x,g_y] \gets gradient(B)$.  // Gradient operation, using central differencing scheme.
                \State $V_x \gets g_x \exp(-|g_x|)$  // Local change in brightness (stimuli) along x axis.
\State $V_y \gets g_y \exp(-|g_y|)$ //Local change in brightness (stimuli) along y axis.
\State $V \gets \sqrt{V_x^2+V_y^2}$  //Local stimuli map.
                \EndProcedure
        \end{algorithmic}
\end{algorithm}
\footnotetext{The scaling factor $k$ is set as 1 in our experiments. The sample implementation of the code in MATLAB can be obtained by contacting the authors.}

Algorithm 1 provides the formal summary of the proposed  feature enhancement process based on local stimuli weighting. Figure 2 shows a step wise graphical illustration of algorithm 1 and indicates the proposed methods ability to detect and amplify edge signals in step 3-4. The proposed method retrieve edges by computing the local stimuli experienced at each pixel location with the help of the amount of brightness perceived to form the image. The local stimuli is computed as the two dimensional variations in perceived brightness. The natural variability in computed brightness is observed to be less; at the same time edge information is sufficient to compute the responses. Also exponential weights are introduced to the filtering scheme, which further helps in the suppression of intra-region intensity variability.\begin{figure}[ht]
\centering
\includegraphics[width=80mm]{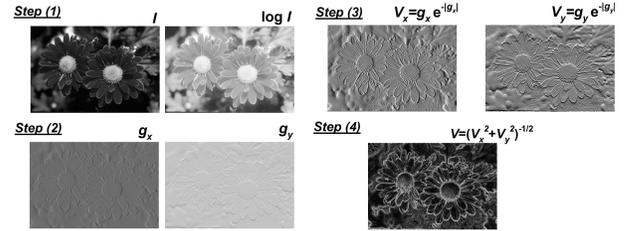}
\caption{A  graphical illustration on the working of the proposed edge detection method. \ }
\end{figure}

 Figure 3 illustrates the comparison of the proposed method for the amount of underlying structural information extracted with that of the existing methods such as Sobel, SIS, Kirsch, and Prewitt \cite{8} for a sample face image. The images visually illustrate that the proposed method produces meaningful edge features compared to other methods, where the binary edges are obtained using the adaptive Ostu's threshold technique\cite{19}. Proposed method shows least spurious structures with much better distinction from corresponding background.
Edge responses computed for a set of images with diverse imaging conditions from Berkeley Segmentation Dataset\footnote{http://www.eecs.berkeley.edu/Research/Projects/CS/vision/bsds/} is displayed in Fig. 4. 

\begin{figure}[hbtp]
\center
\subfigure[Sobel]{
  \includegraphics[width=.65in]{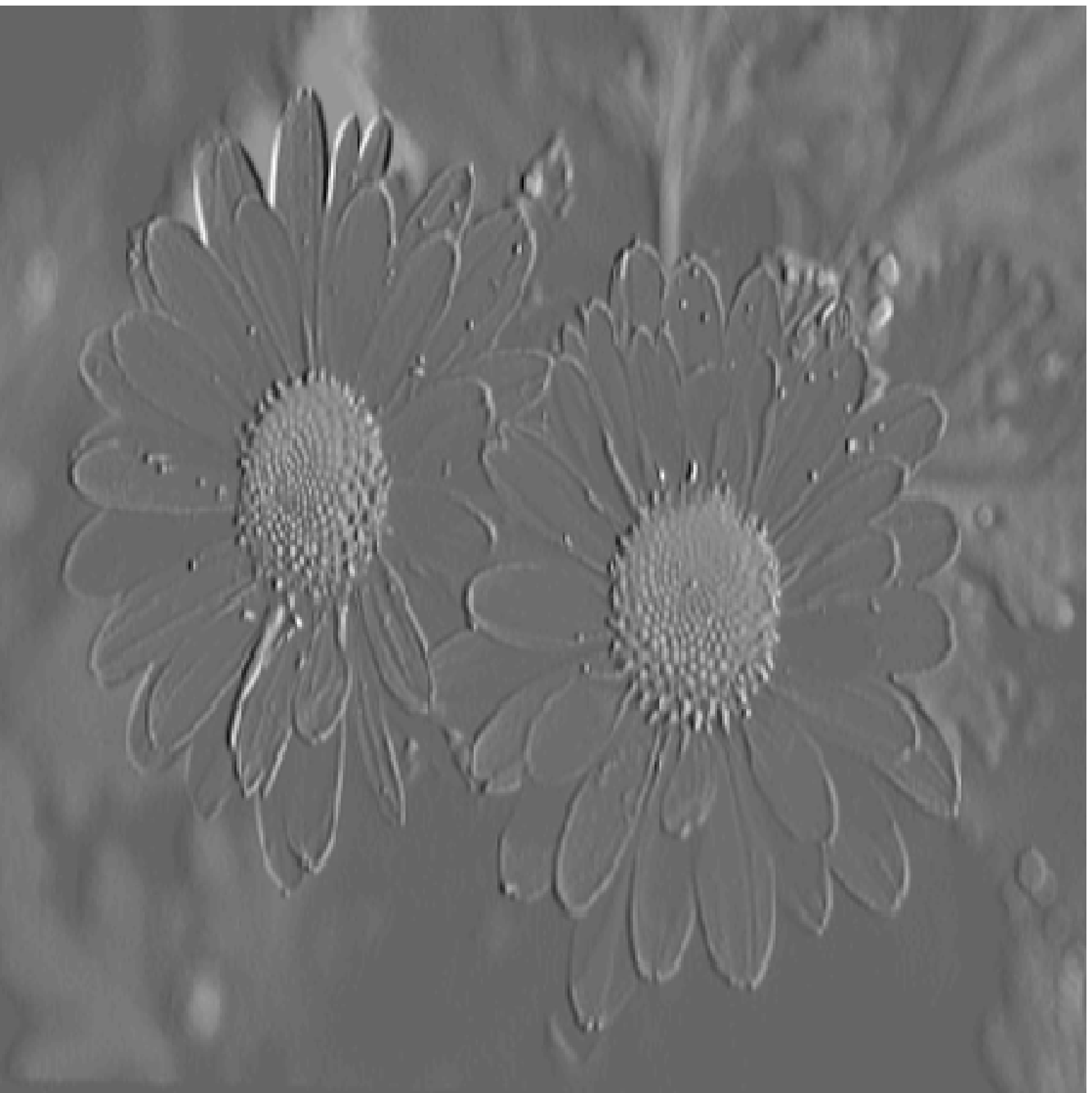}
   \label{fig5:subfig1}
   }\hspace*{-0.9em}
\subfigure[SIS]{
  \includegraphics[width=.65in]{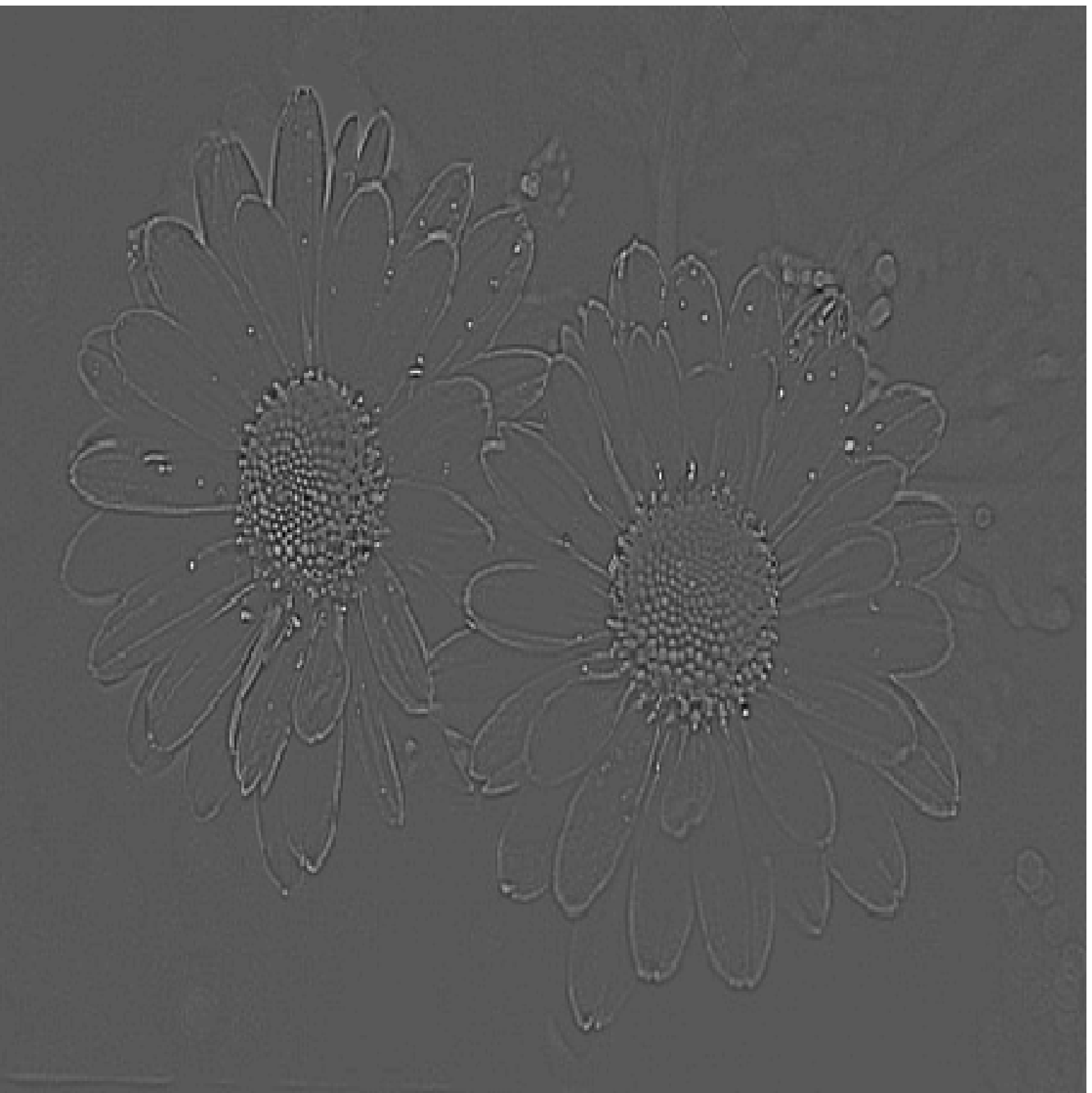}
   \label{fig5:subfig2}
   }\hspace*{-0.9em}
\subfigure[Kirsch]{
  \includegraphics[width=.65in]{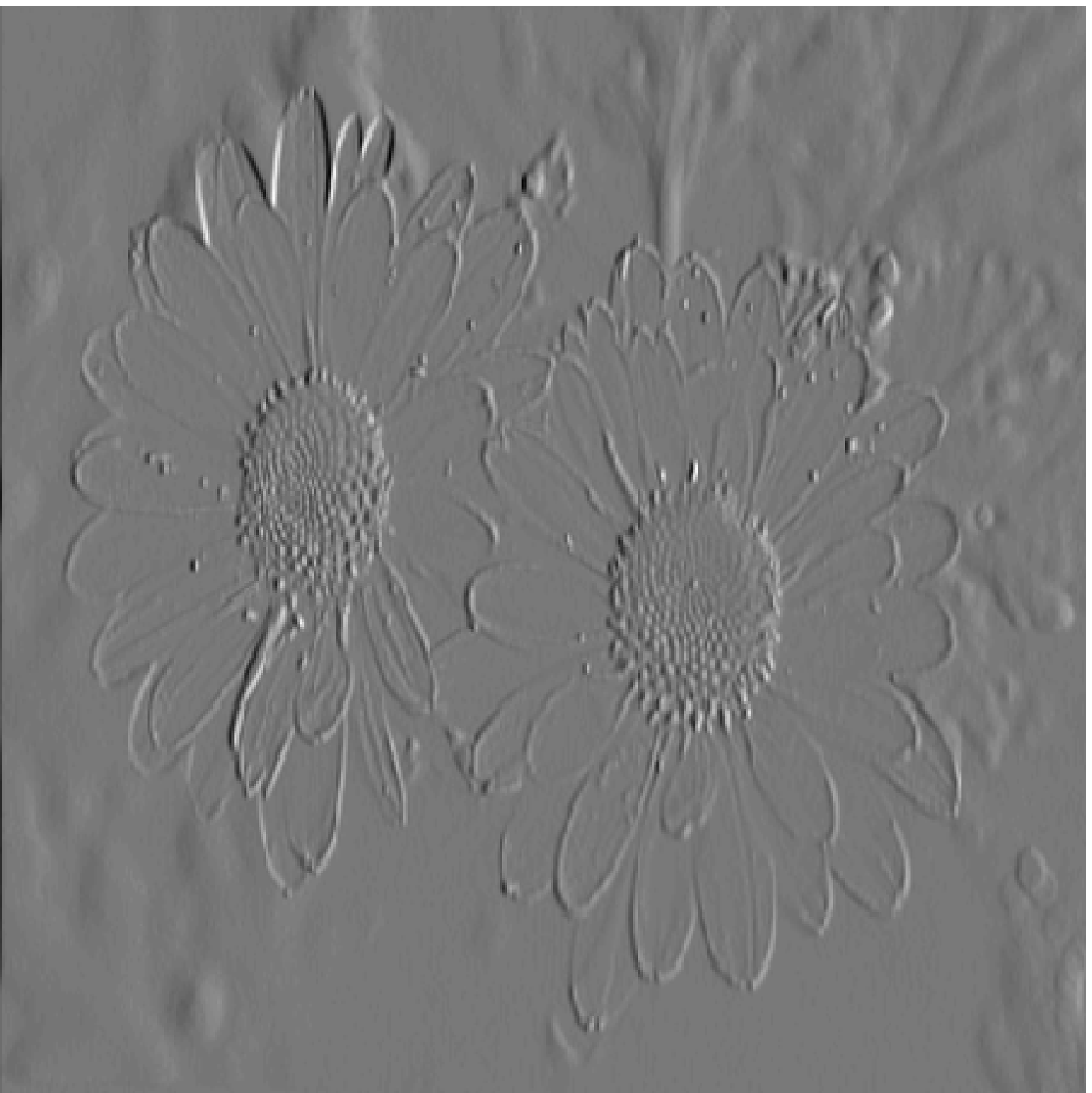}
   \label{fig5:subfig3}
   }\hspace*{-0.9em}
\subfigure[Prewitt]{
  \includegraphics[width=.65in]{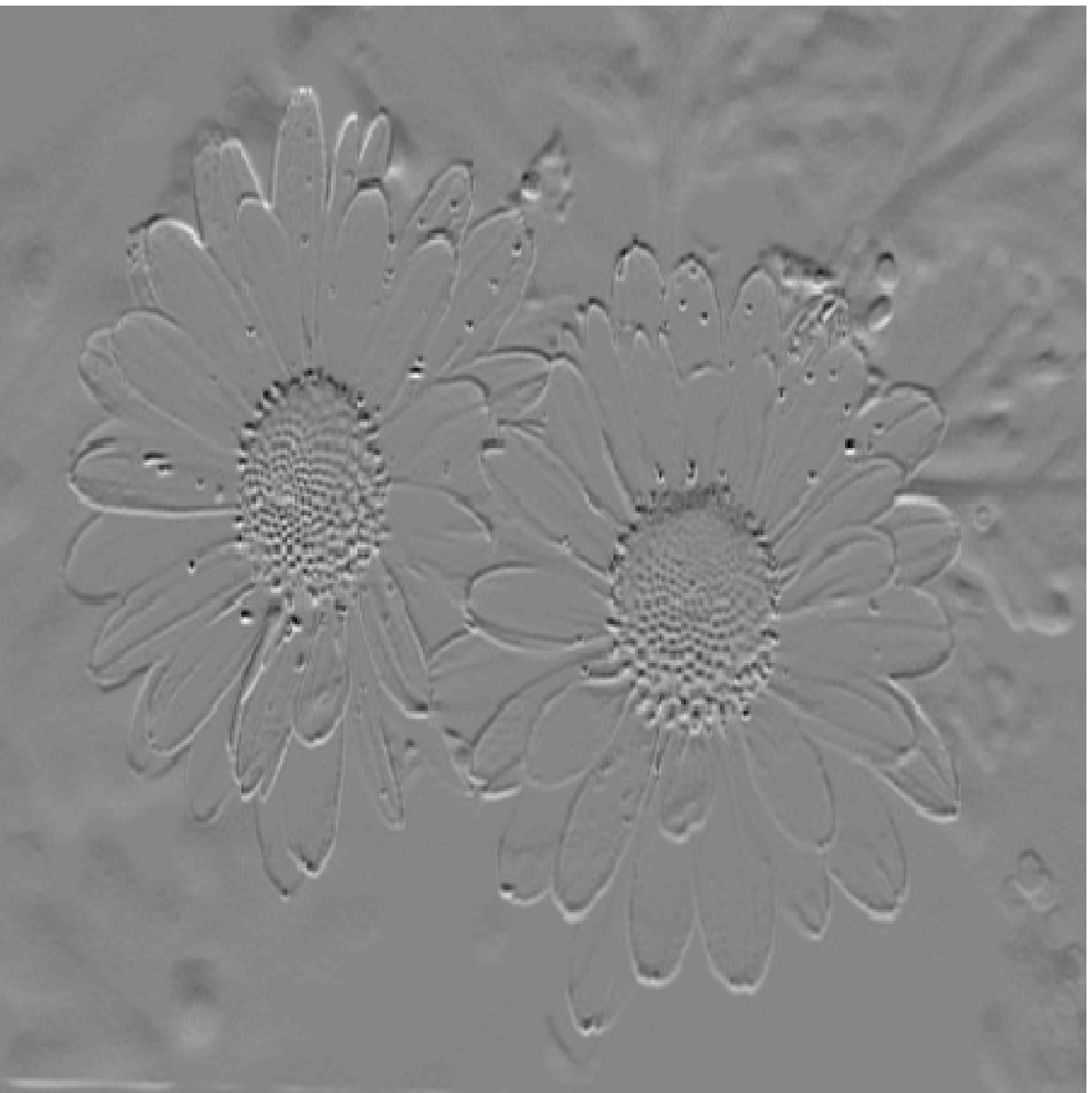}
   \label{fig5:subfig4}
   }\hspace*{-0.9em}
\subfigure[Proposed]{
  \includegraphics[width=.65in]{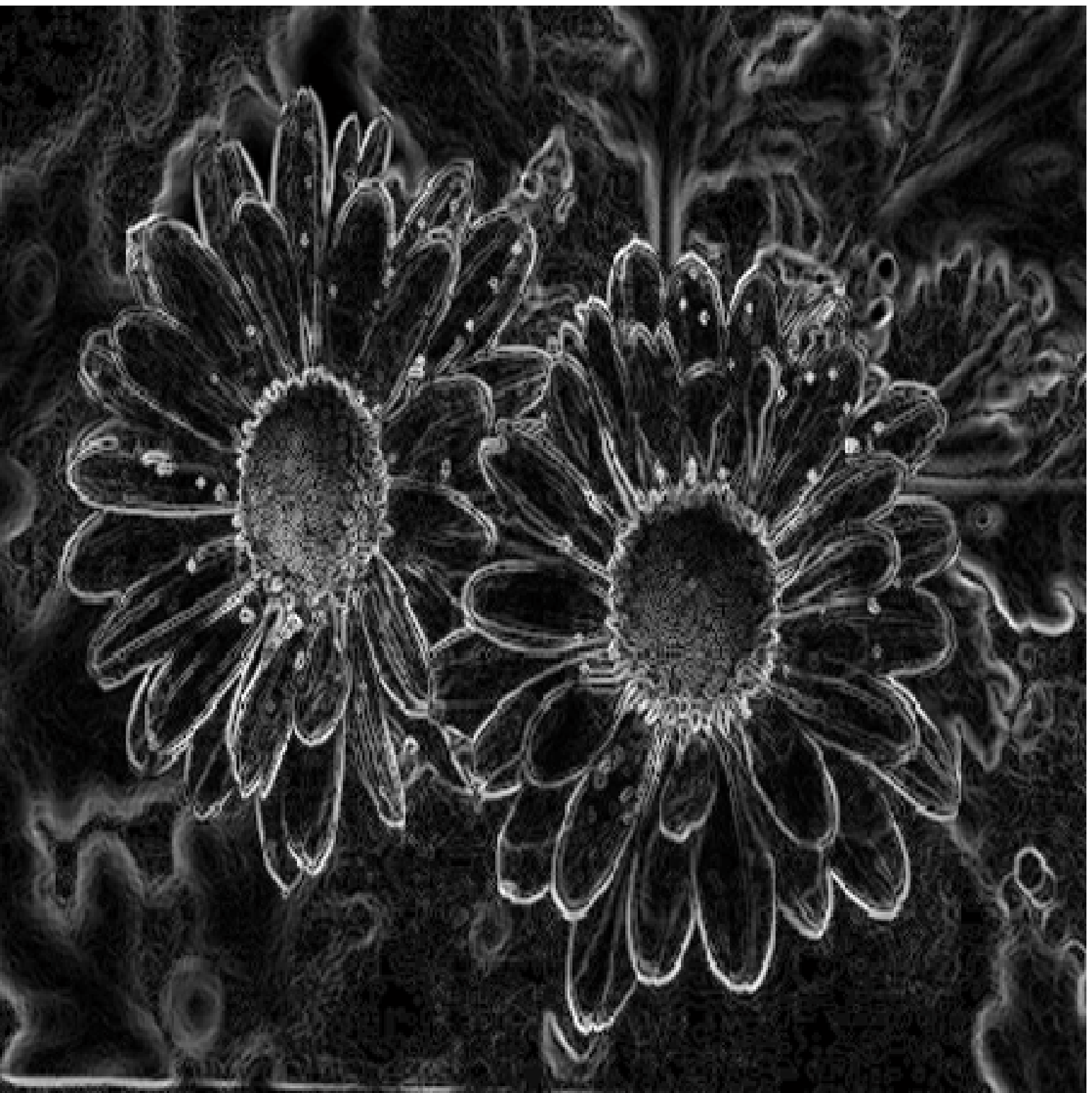}
   \label{fig5:subfig5}
   }
\subfigure[Sobel]{
  \includegraphics[width=.65in]{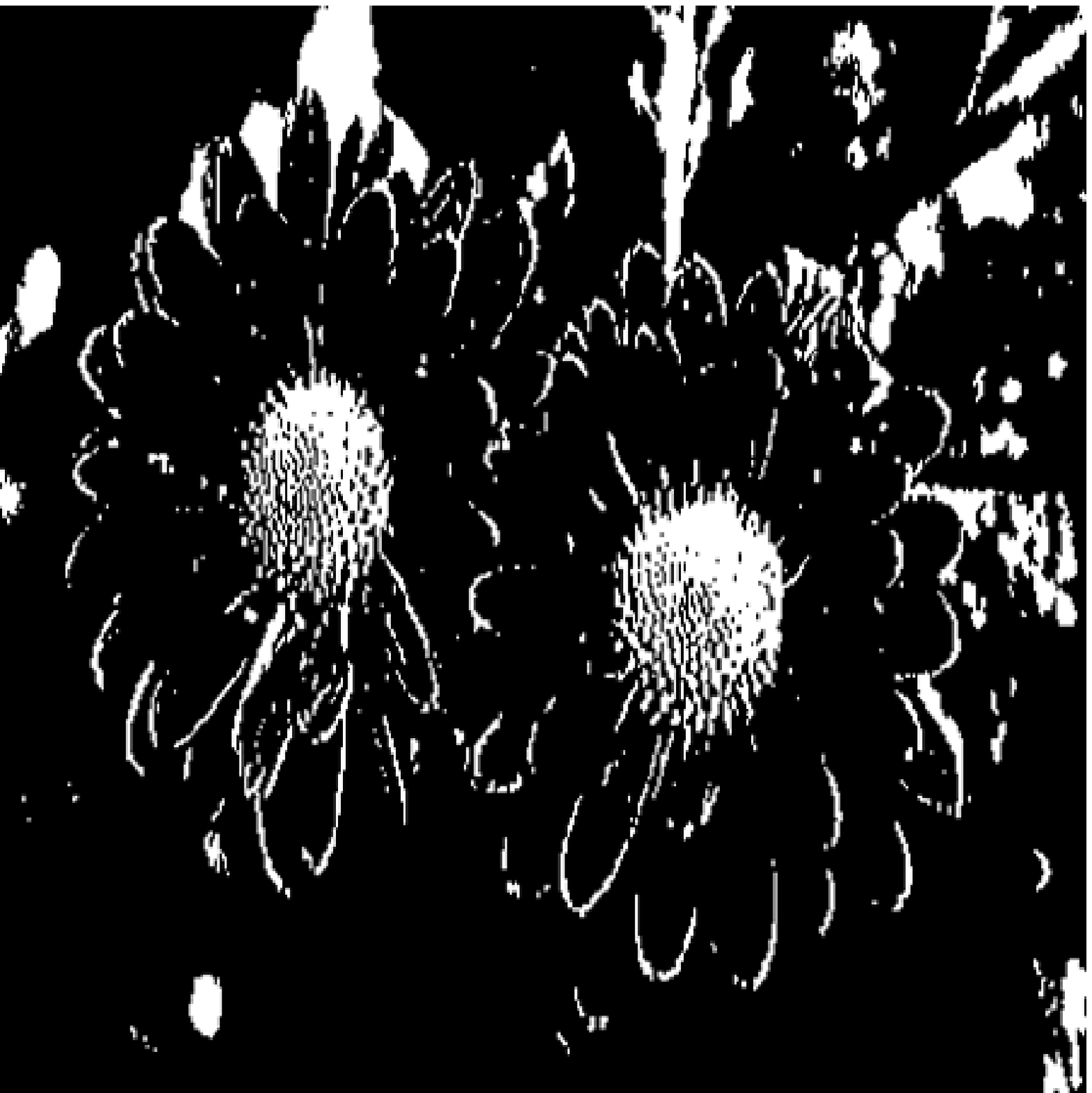}
   \label{fig5:subfig6}
   }\hspace*{-0.9em}
\subfigure[SIS]{
  \includegraphics[width=.65in]{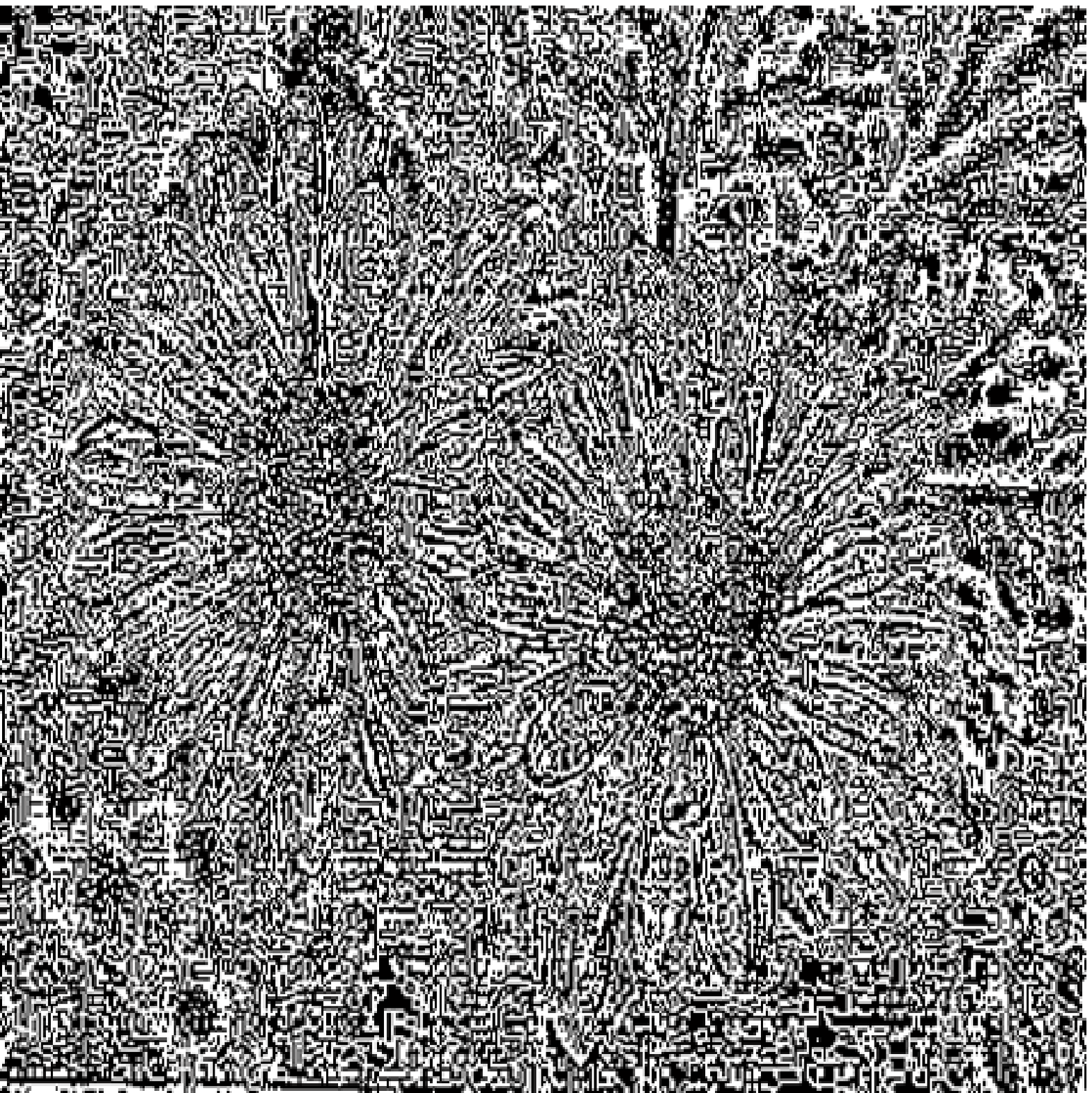}
   \label{fig5:subfig7}
   }\hspace*{-0.9em}
\subfigure[Kirsch]{
  \includegraphics[width=.65in]{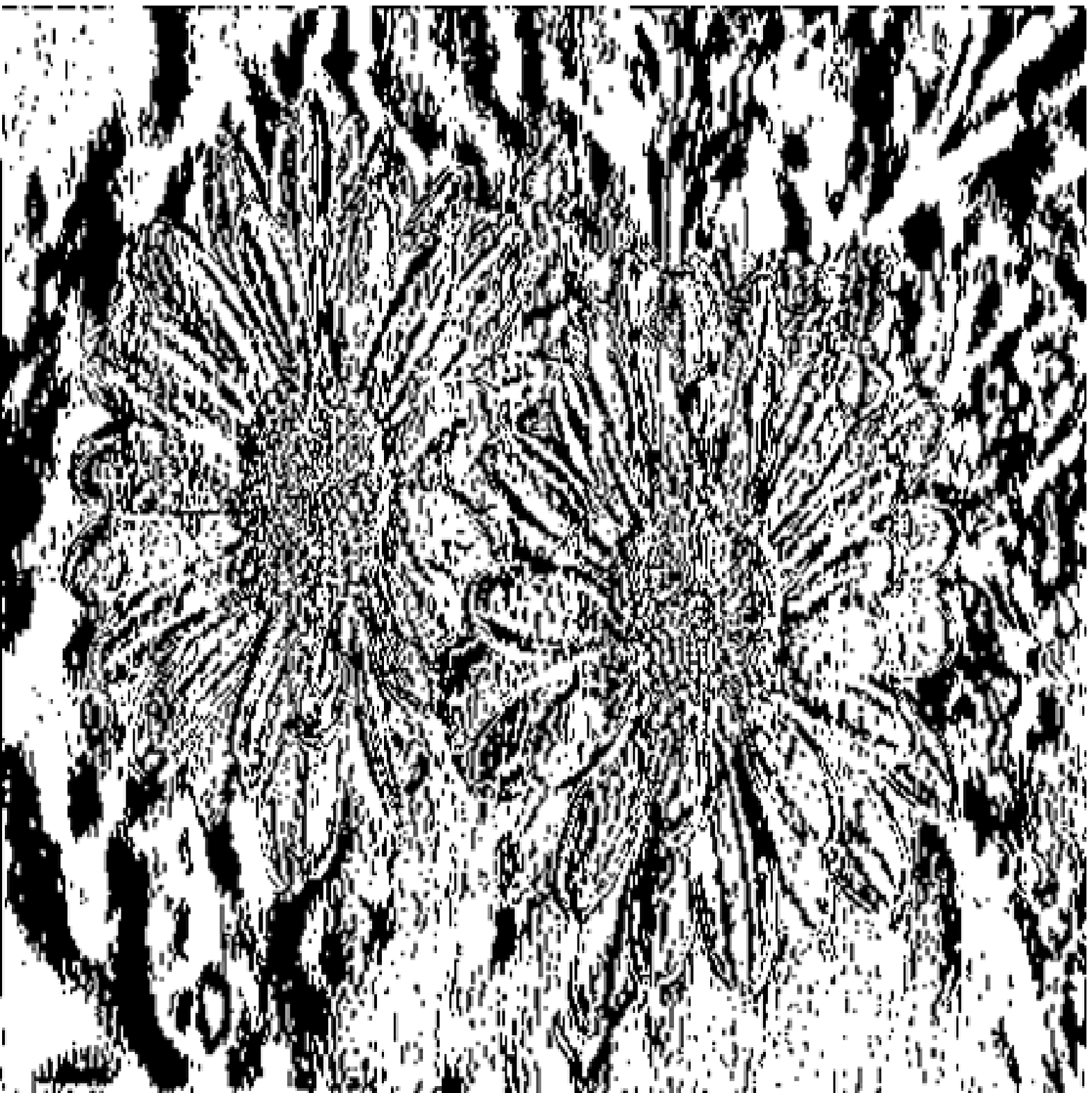}
   \label{fig5:subfig8}
   }\hspace*{-0.9em}
\subfigure[Prewitt]{
  \includegraphics[width=.65in]{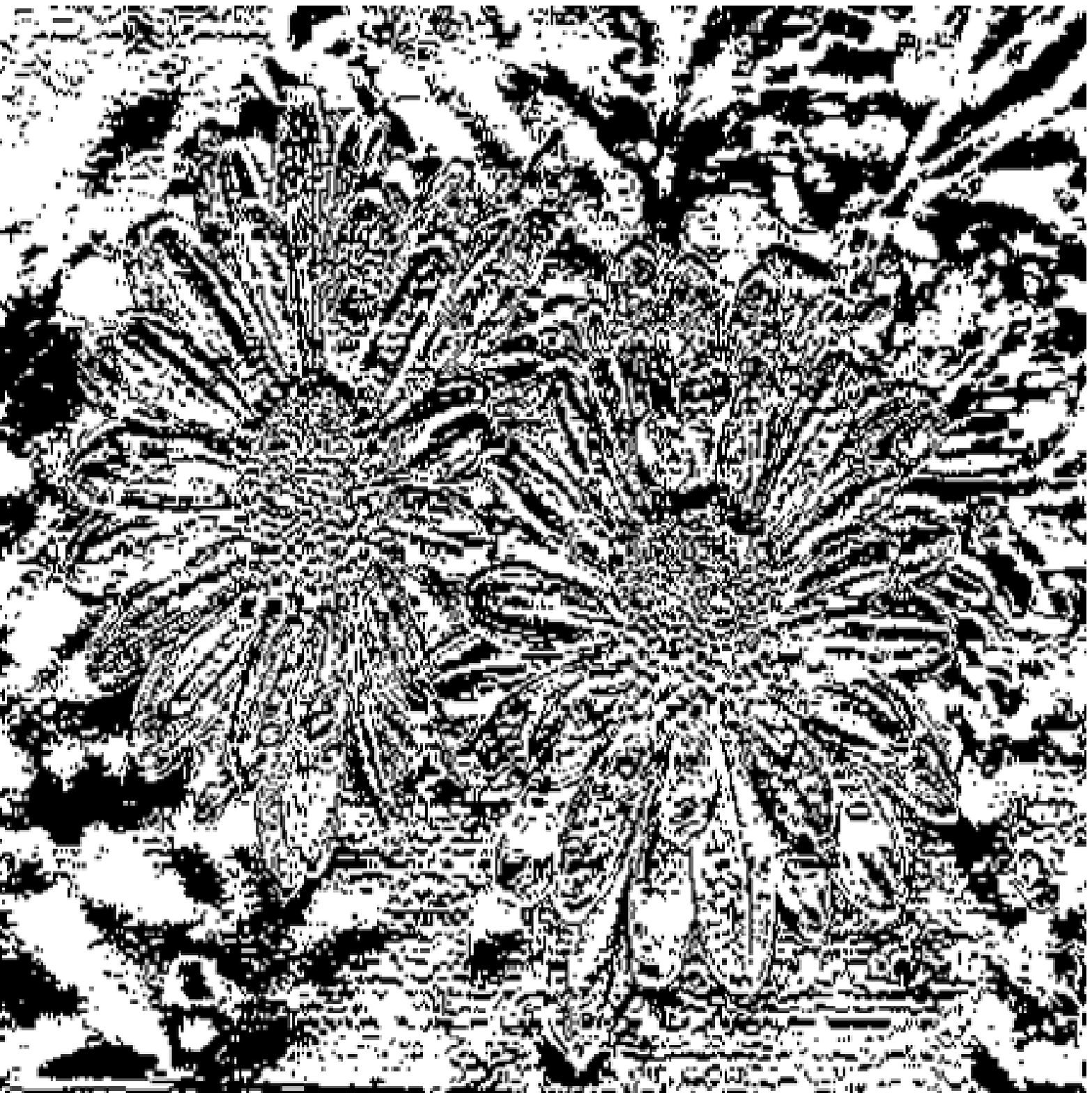}
   \label{fig5:subfig9}
   }\hspace*{-0.9em}
\subfigure[Proposed]{
  \includegraphics[width=.65in]{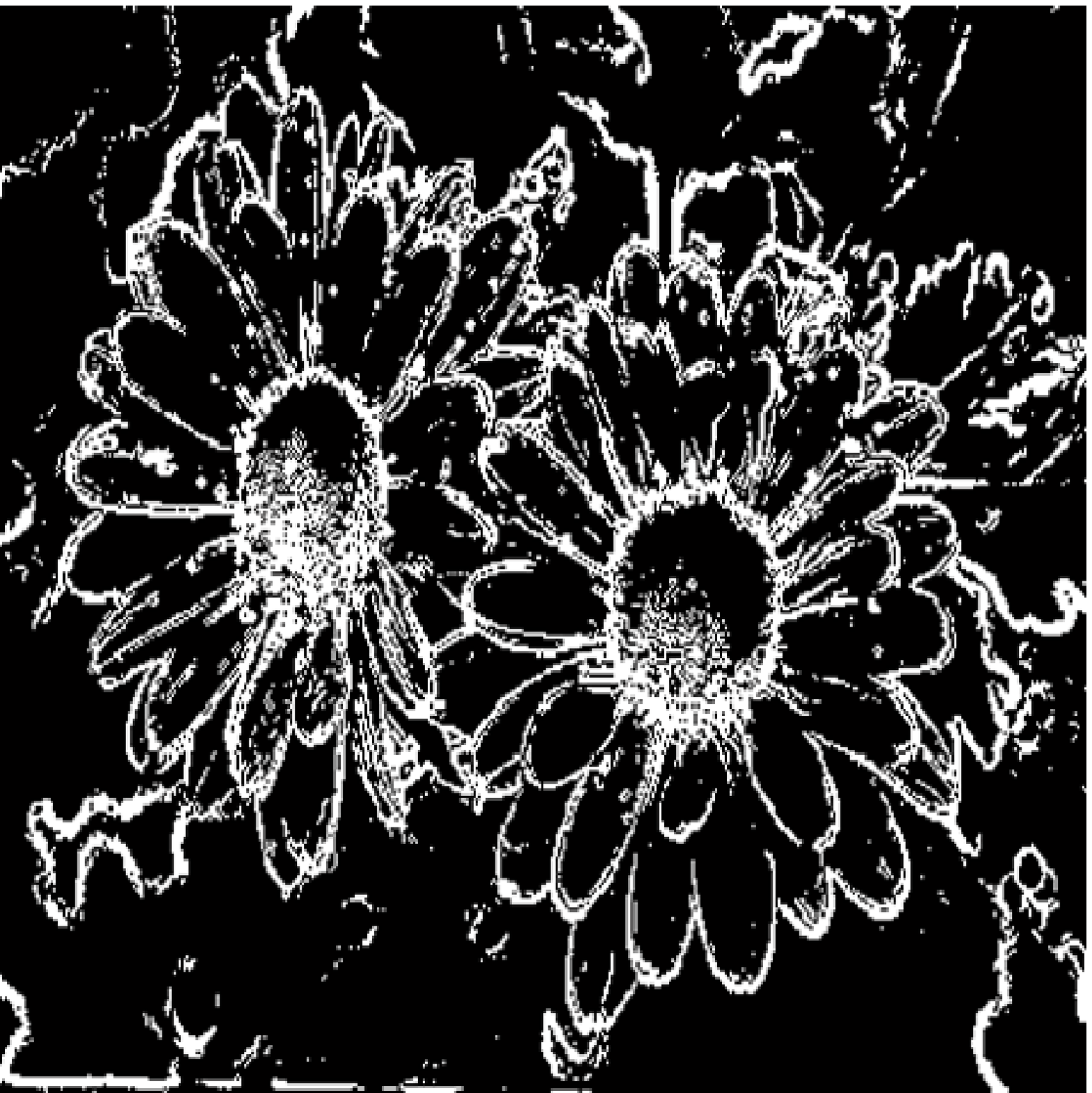}
   \label{fig5:subfig10}
   }
\caption{Detected edges and binary maps computed for visualizing edge response (a-e) and the underlying structural details (f-j) of original image extracted using various methods namely Sobel, SIS, Kirsch and Prewitt, in comparison to the proposed method}
\label{figure5}
\end{figure}

\begin{figure}[hbtp]
\center
\subfigure[]{
  \includegraphics[width=.65in]{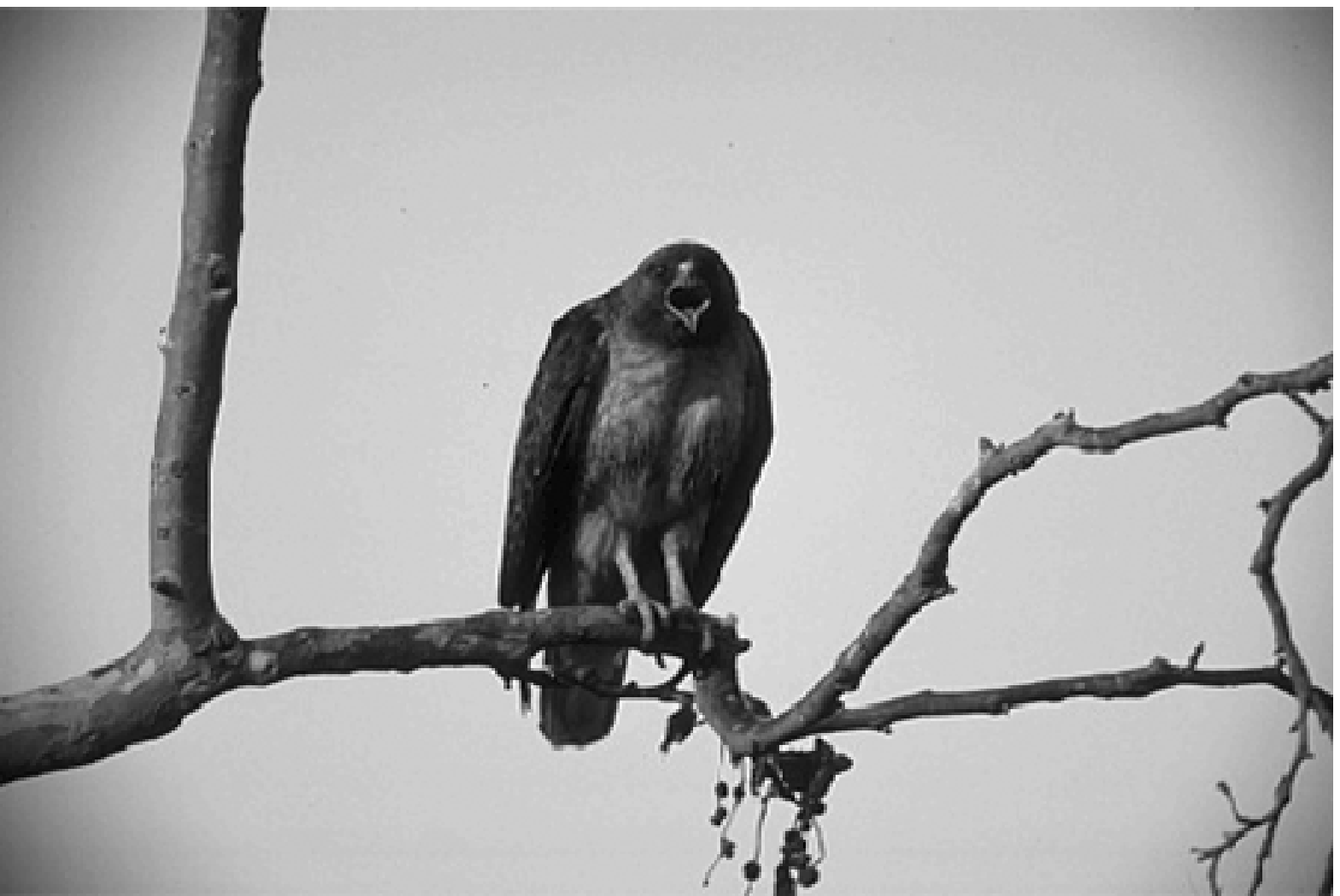}
   \label{fig6:subfig1}
   }\hspace*{-0.9em}
\subfigure[]{
  \includegraphics[width=.65in]{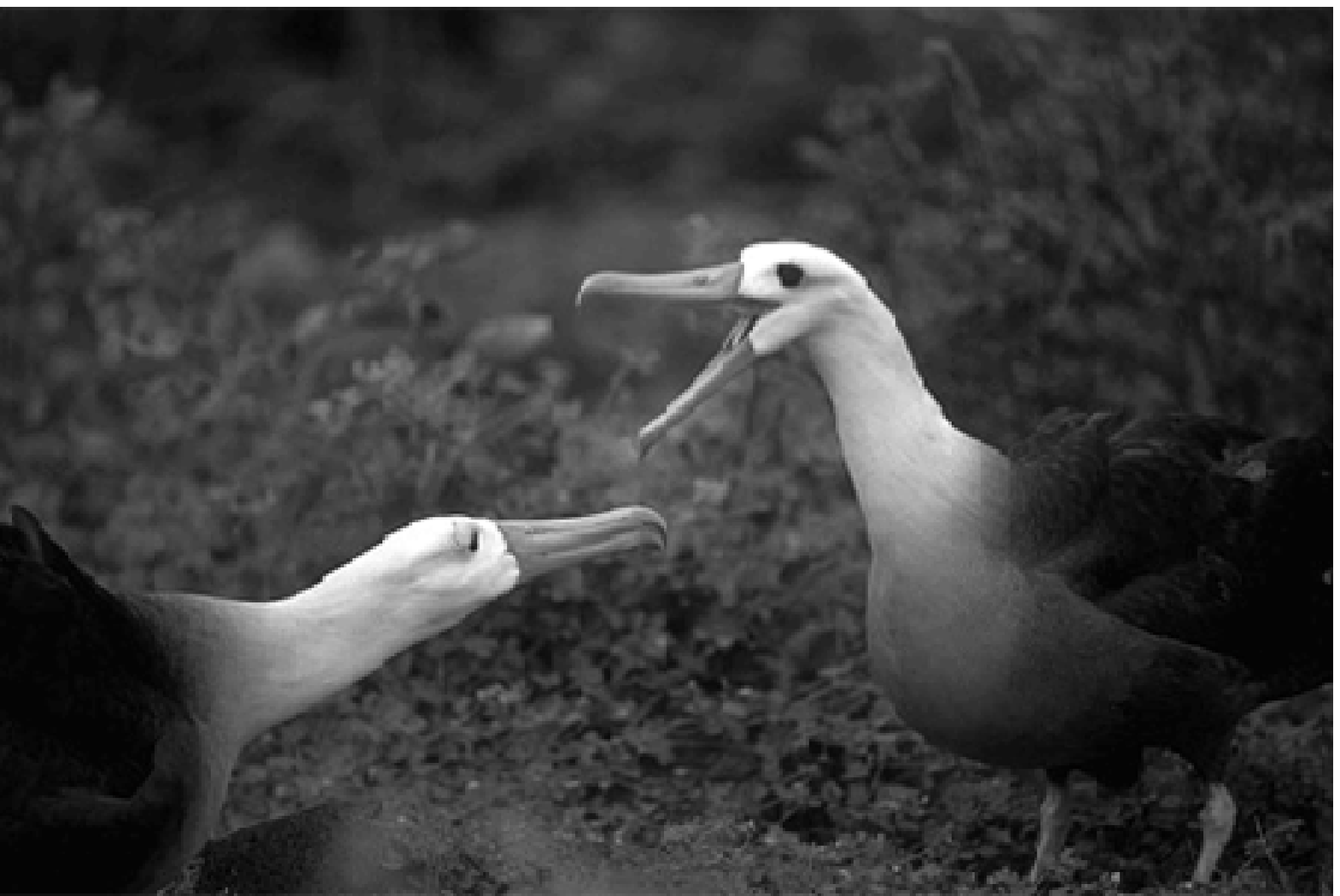}
   \label{fig6:subfig3}
   }\hspace*{-0.9em}
   \subfigure[]{
  \includegraphics[width=.65in]{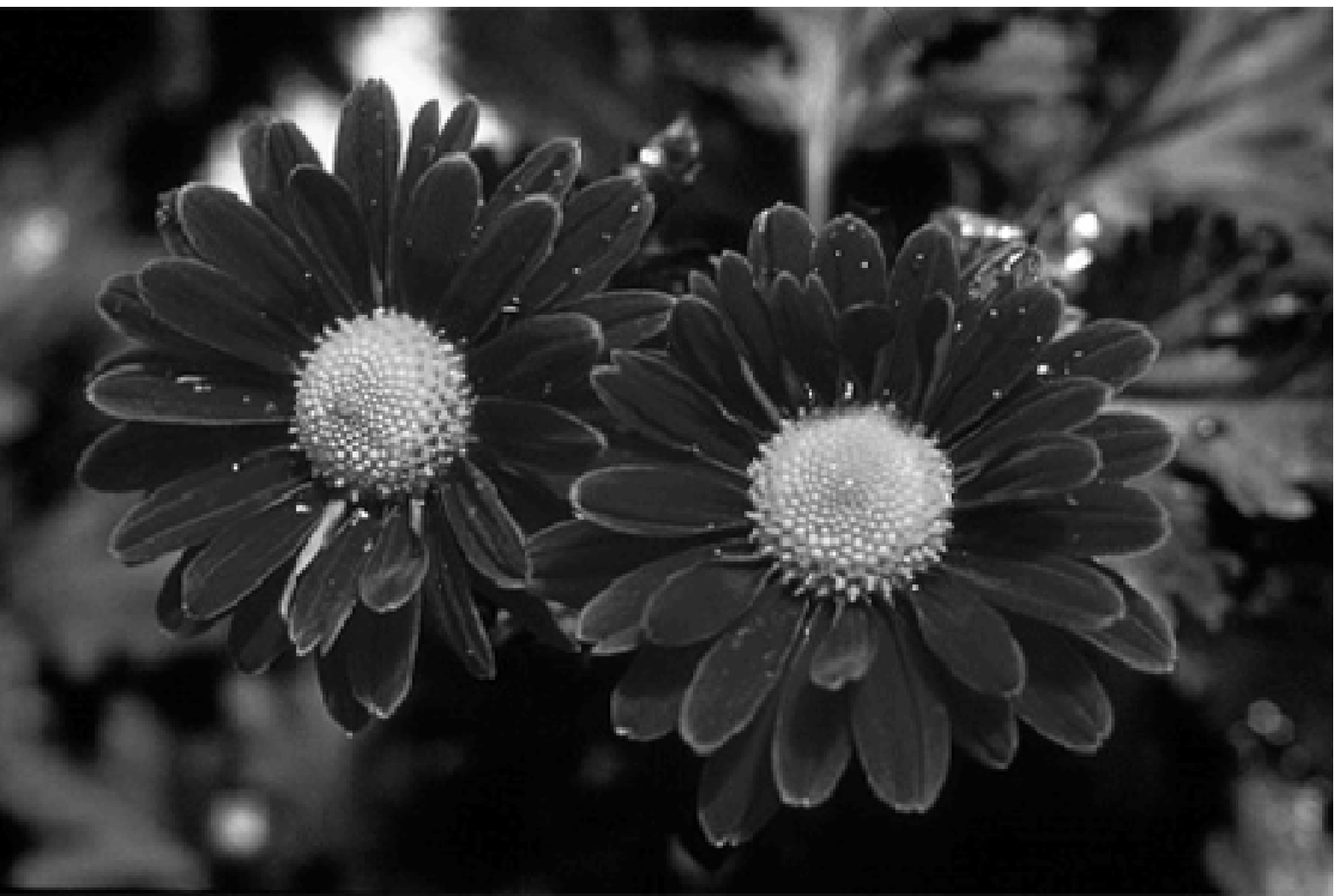}
   \label{fig6:subfig5}
   }\hspace*{-0.9em}
   \subfigure[]{
  \includegraphics[width=.65in]{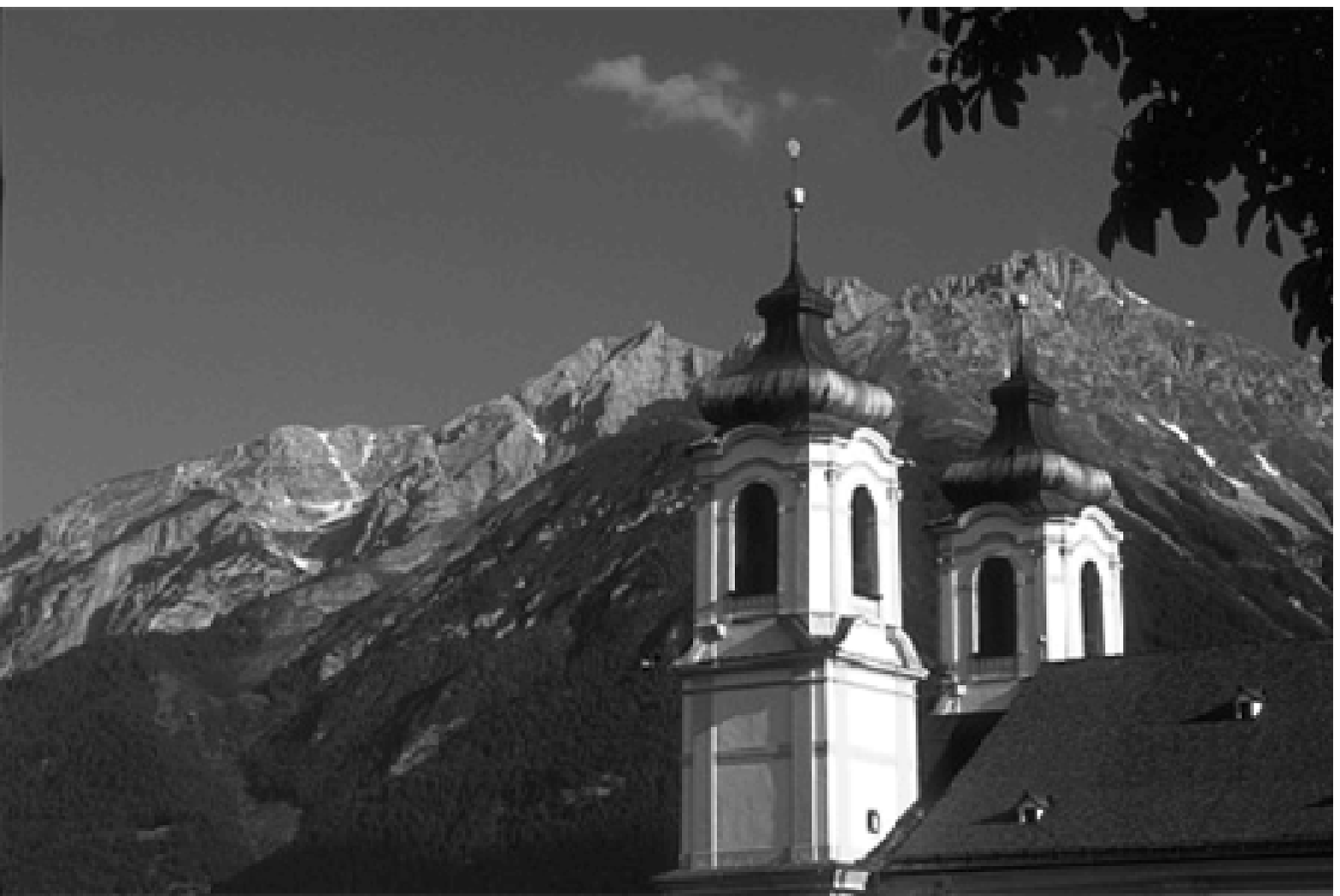}
   \label{fig6:subfig7}
   }\hspace*{-0.9em}
   \subfigure[]{
  \includegraphics[width=.65in]{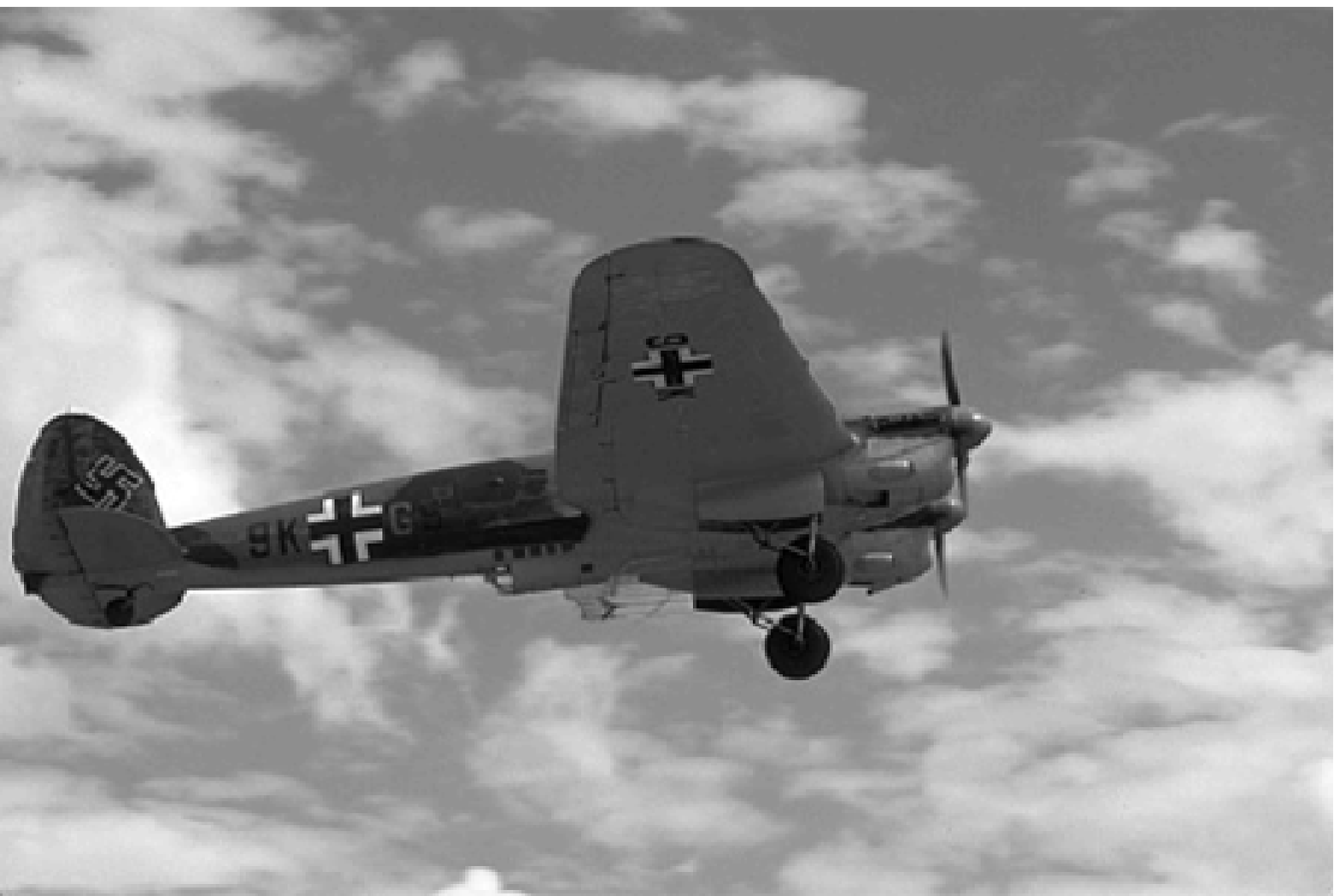}
   \label{fig6:subfig9}
   }
   \subfigure[]{
  \includegraphics[width=.65in]{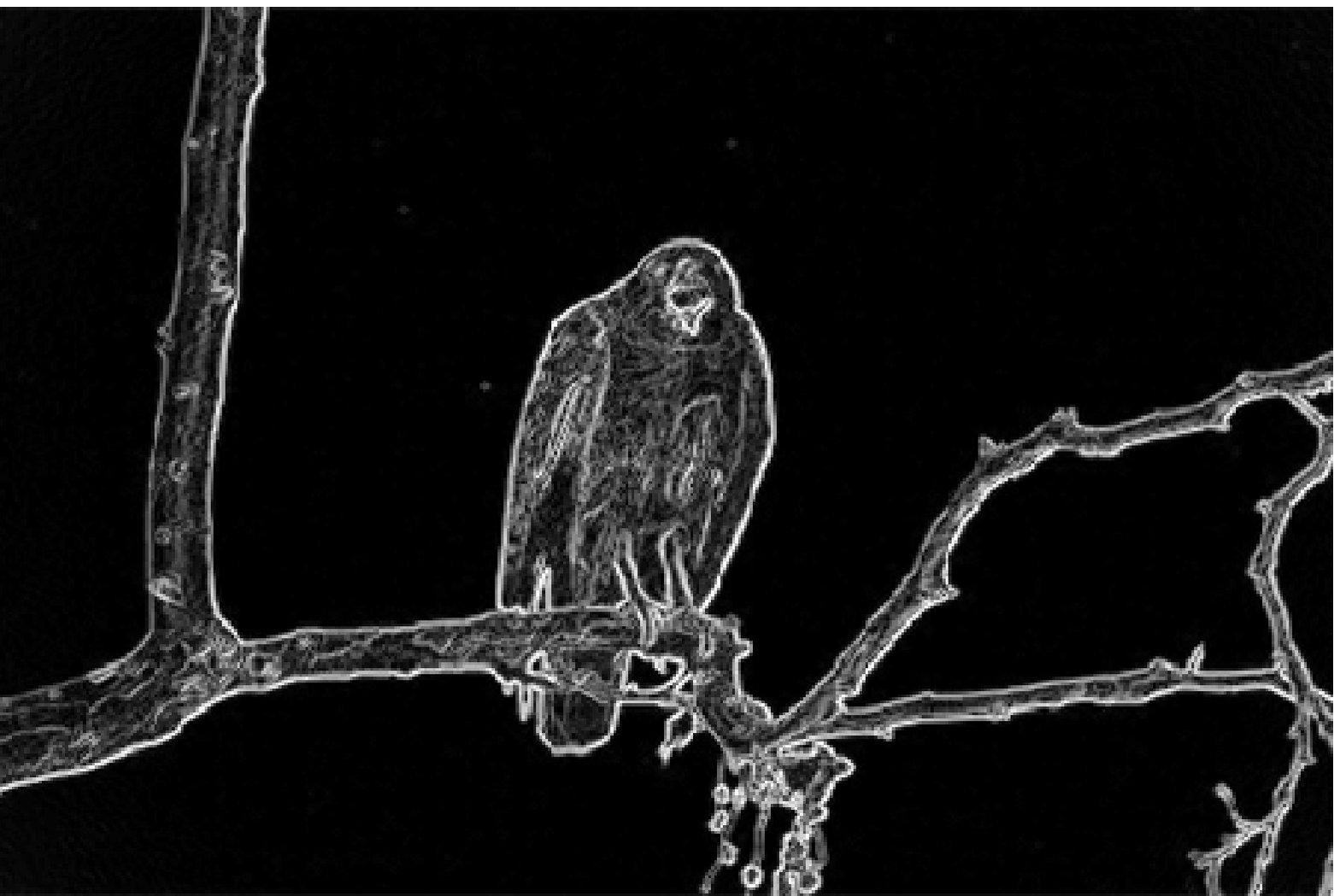}
   \label{fig6:subfig2}
   }\hspace*{-0.9em}
\subfigure[]{
  \includegraphics[width=.65in]{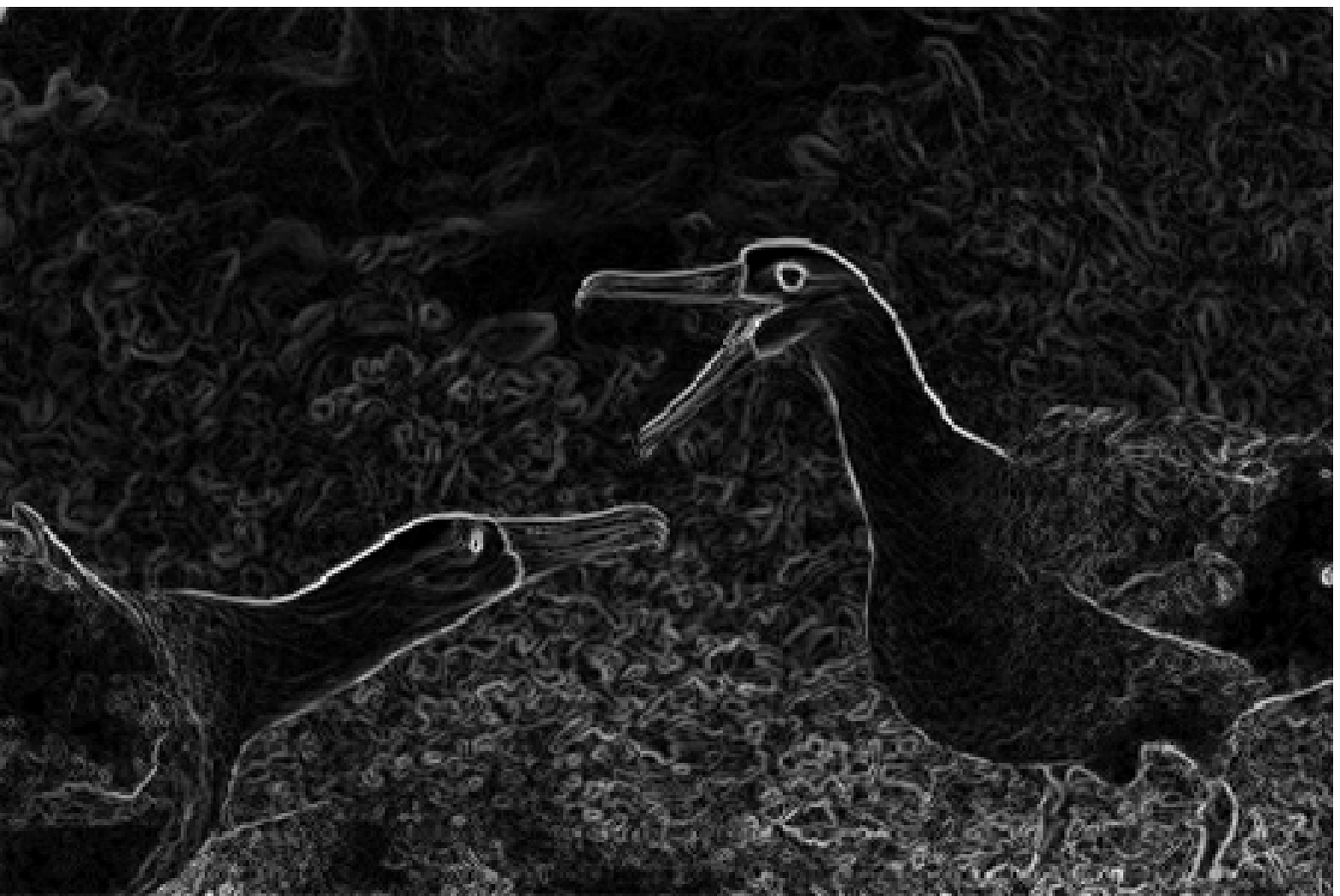}
   \label{fig6:subfig4}
   }\hspace*{-0.9em}
\subfigure[]{
  \includegraphics[width=.65in]{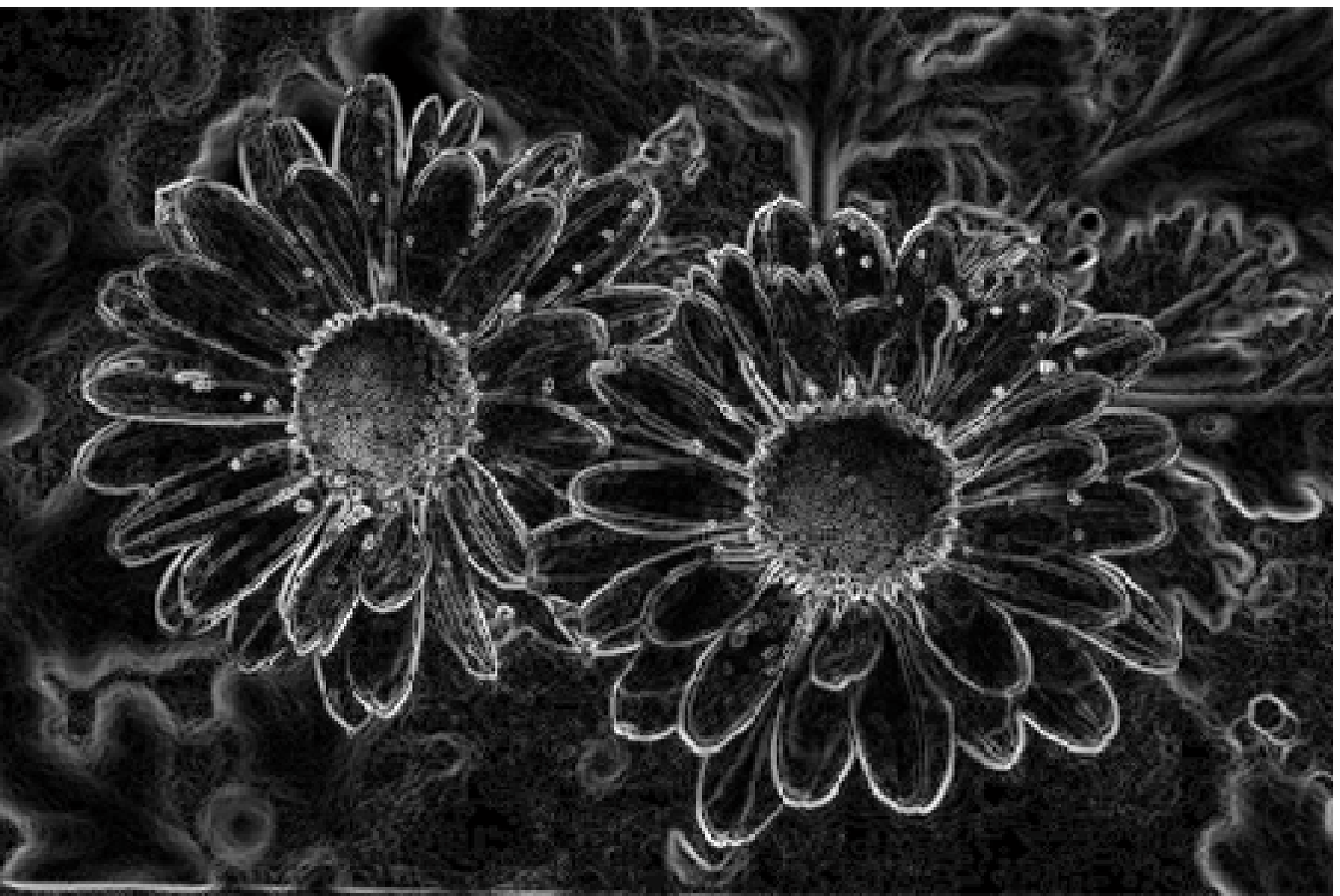}
   \label{fig6:subfig6}
   }\hspace*{-0.9em}
\subfigure[]{
  \includegraphics[width=.65in]{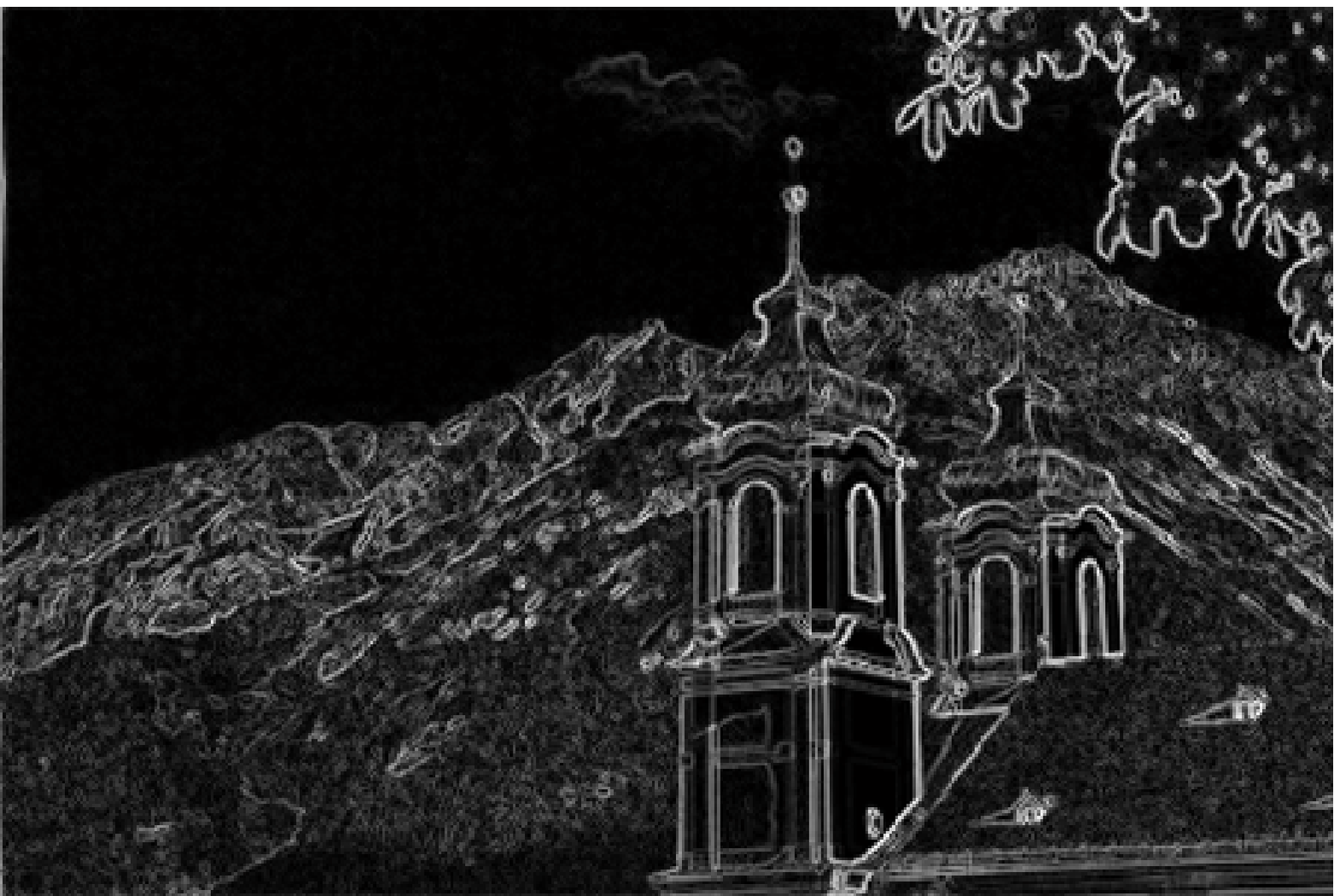}
   \label{fig6:subfig8}
   }\hspace*{-0.9em}
\subfigure[]{
  \includegraphics[width=.65in]{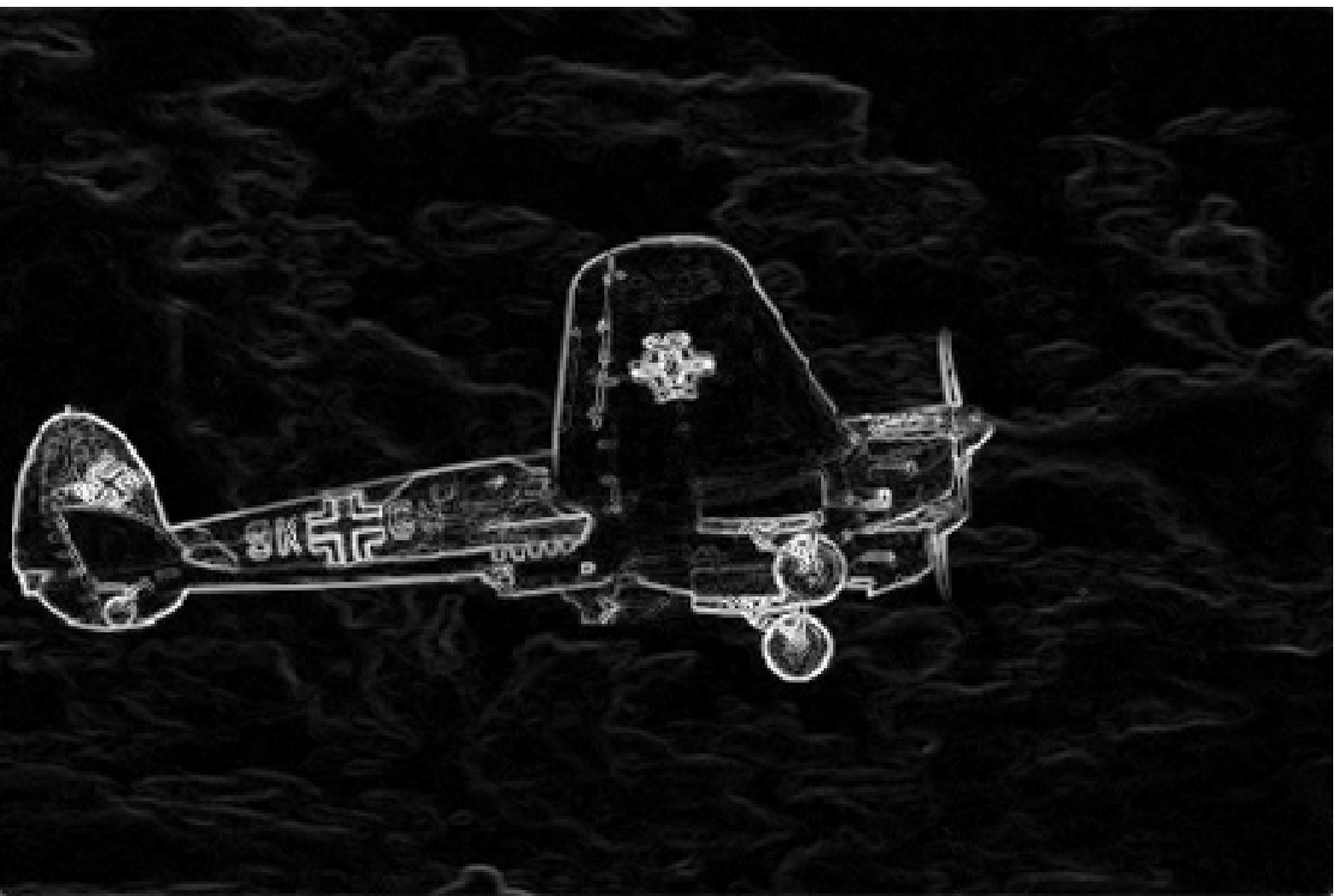}
   \label{fig6:subfig10}
   }
\caption{Edge responses computed for natural scenes  shared by Berkeley Segmentation Dataset, with original images (a-e) and corresponding edge responses (f-j).}
\label{figure7}
\end{figure}

\section{Experimental Results}

For better understanding, the responses to edge computed for classical primal sketch model\footnote{Primal Sketch model results shown are based on [2], "theory of edge detection" by Marr, where an edge is defined as the zero crossing along region transitions. And to display the fundamental model of edge detection, we use second order operation represented by Laplacian operator.} and proposed scheme for various shapes are shown in Table \ref{table1},\ref{table2},\ref{table3}. Table \ref{table1} illustrates the edges computed for varying gaussian noise  ($\sigma_{nv}^2$) in the order of 0\%, 5\%, 10\%, and 20\% of the maximum of image intensities for step-edge. From results, primal sketch model based scheme fails to detect  edges beyond 5\% increase in variability, whereas the proposed method is proves to be capable of detecting edges even at 20\% intensity variability. This indicates the proposed method''s capability in extracting maximum number of sketchable edges with minimal unsketchable details. 

Table \ref{table2} illustrates the response on Gaussian shaped intensity changes with increasing intensity variability. Results depicts that the primal sketch model without smoothing fails even with slight increase in variability. Whereas, the proposed method shows a good tolerance against increasing intensity variability with a comparable edge localization. Table \ref{table3} depicts spatial edge response for a ramp shape, where it can be observed that variability tolerance for method schemes are very poor, whereas the proposed method is shown to be more robust to pixel noise.

\begin{table}[h!]
\label{table1}
        \begin{center}
                \caption{Comparison of Edge Responses computed using a step shape for varying percentile standard deviations in reference to the maximum intensity value}
                \begin{tabular}{ | p{0.7cm} | p{2.cm} | p{2.cm} | p{2.cm} |}
                        \hline
                        $\sigma_{nv}^2$(\%) & Original & Primal Sketch Model & Proposed \\ \hline
                 0 &    \includegraphics[width=0.5in]{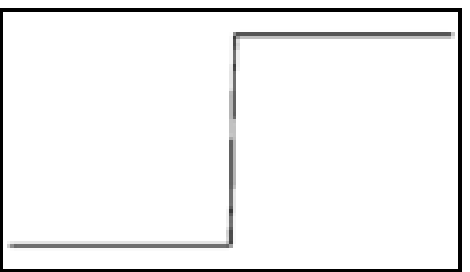} & 
                \includegraphics[width=0.5in]{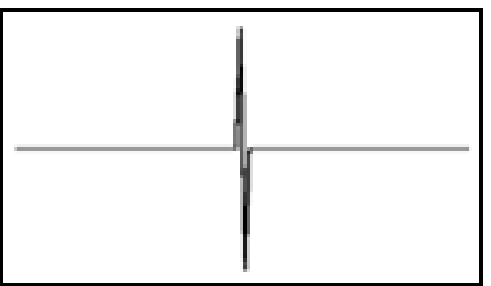}
                        & 
                \includegraphics[width=0.5in]{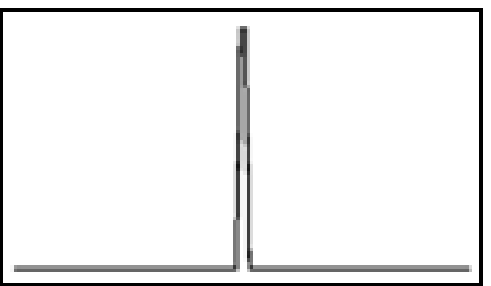} 
                         \\ \hline
                         
                                5 &     \includegraphics[width=0.5in]{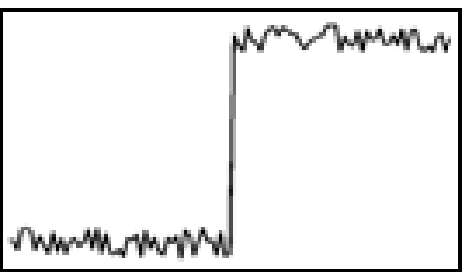} & 
                                 \includegraphics[width=0.5in]{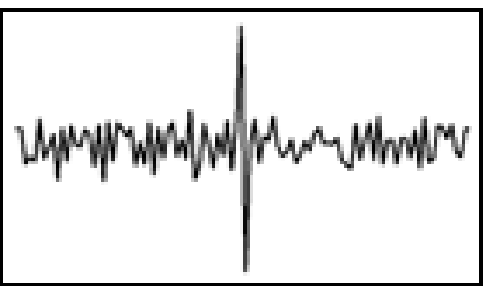}
                                 & 
                                 \includegraphics[width=0.5in]{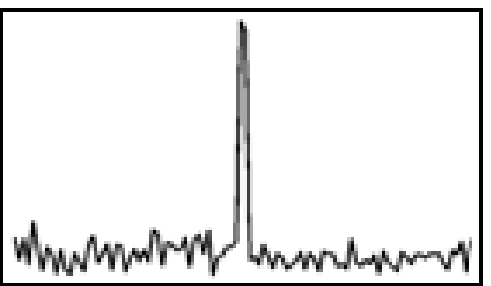}   \\ \hline
                                 
                                10 &    \includegraphics[width=0.5in]{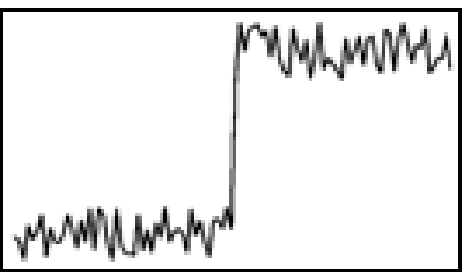} & 
                                  \includegraphics[width=0.5in]{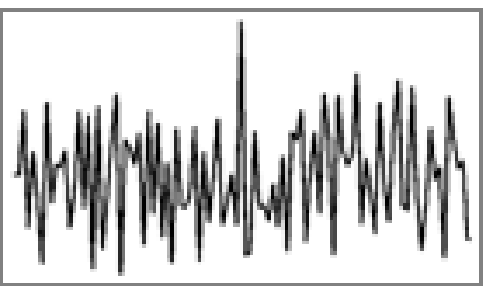}
                                  & 
                                  \includegraphics[width=0.5in]{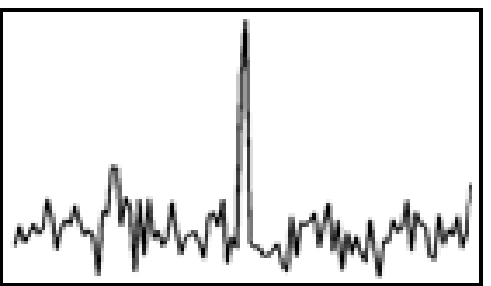}  \\ \hline
                                  
                                   20 & \includegraphics[width=0.5in]{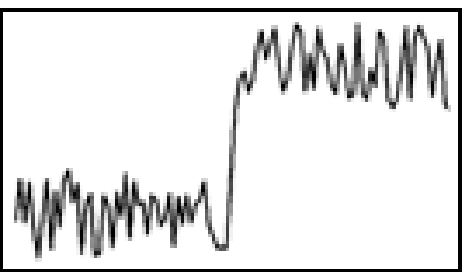} & 
                                   \includegraphics[width=0.5in]{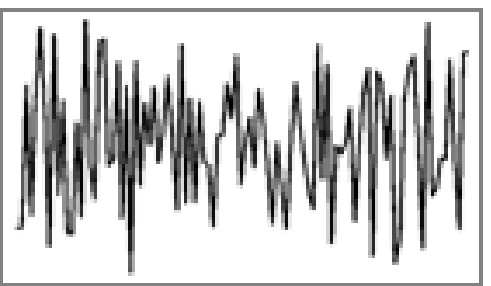}
                                   & 
                                   \includegraphics[width=0.5in]{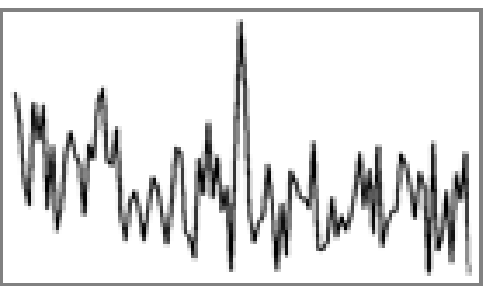}   \\ \hline
                         
                \end{tabular}
                \label{table1}
        \end{center}
\end{table}

\begin{table}[h!]
        \begin{center}
                \caption{Comparison of  Edge Responses computed using a Gaussian shape for varying percentile standard deviations in reference to the maximum intensity value}
                \begin{tabular}{ | p{0.7cm} | p{2.cm} | p{2.cm} | p{2.cm} |}
                        \hline
                        $\sigma_{nv}^2$(\%) & Original & Primal Sketch Model & Proposed \\ \hline
                 0 &    \includegraphics[width=0.5in]{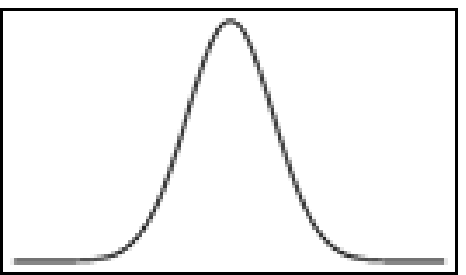} & 
                \includegraphics[width=0.5in]{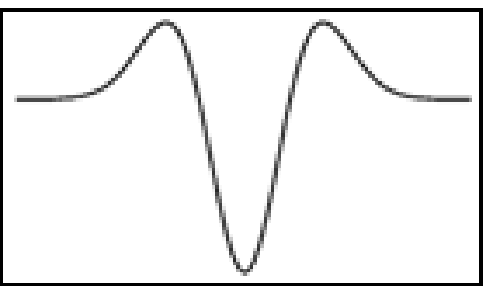}
                        & 
                \includegraphics[width=0.5in]{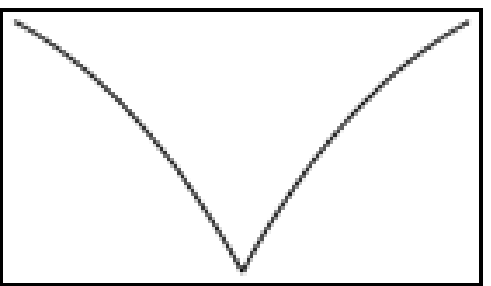} 
                         \\ \hline
                         
                                5 &     \includegraphics[width=0.5in]{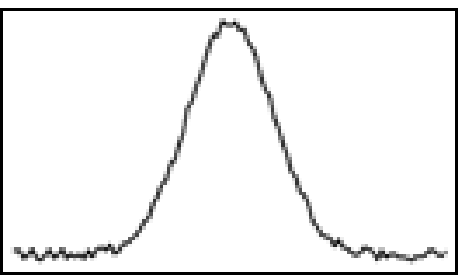} & 
                                 \includegraphics[width=0.5in]{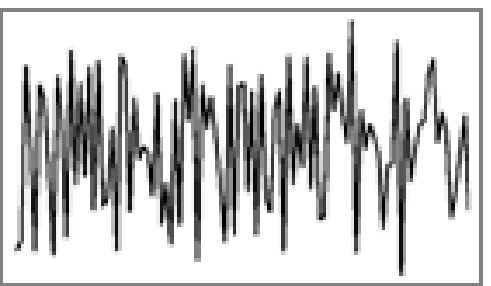}
                                 & 
                                 \includegraphics[width=0.5in]{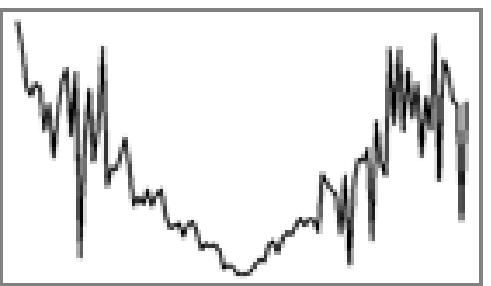}   \\ \hline
                                 
                                10 &    \includegraphics[width=0.5in]{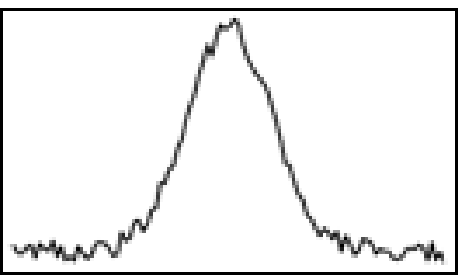} & 
                                  \includegraphics[width=0.5in]{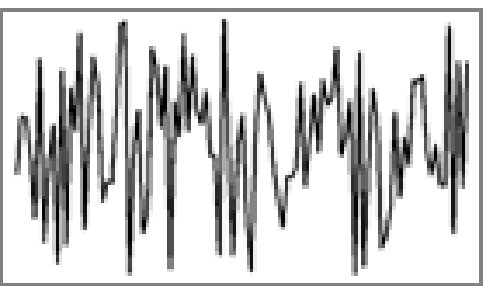}
                                  & 
                                  \includegraphics[width=0.5in]{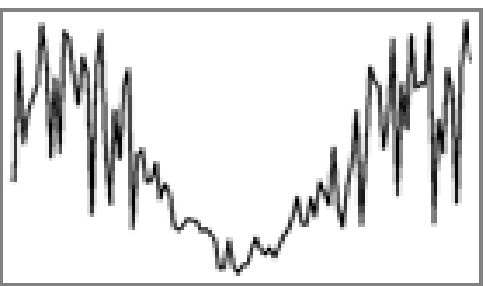}  \\ \hline
                                  
                                   20 & \includegraphics[width=0.5in]{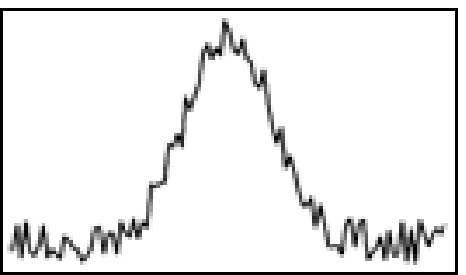} & 
                                   \includegraphics[width=0.5in]{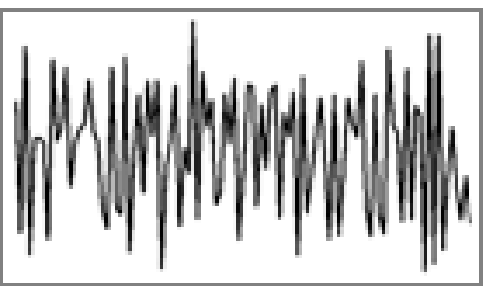}
                                   & 
                                   \includegraphics[width=0.5in]{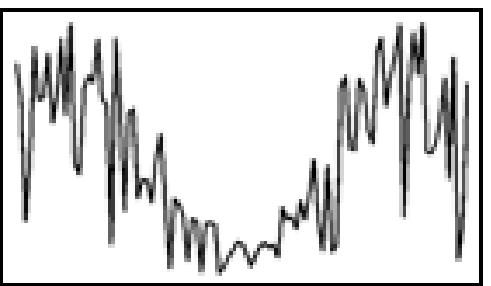}   \\ \hline
                         
                \end{tabular}
                \label{table2}
        \end{center}
\end{table}

\begin{table}[h!]
        \begin{center}
                \caption{Comparison of Edge Responses computed using a ramp shape for varying percentile standard deviations in reference to the maximum intensity value}
                \begin{tabular}{ | p{0.7cm} | p{2.cm} | p{2.cm} | p{2.cm} |}
                        \hline
                        $\sigma_{nv}^2$(\%) & Original & Primal Sketch Model & Proposed \\ \hline
                 0 &    \includegraphics[width=0.5in]{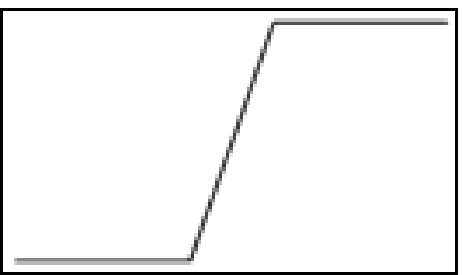} & 
                \includegraphics[width=0.5in]{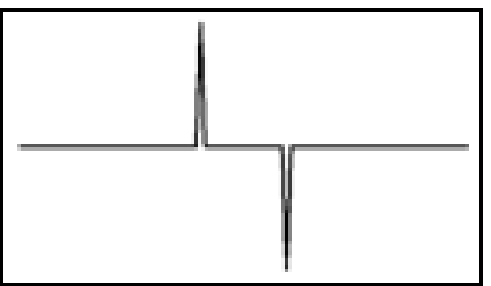}
                        & 
                \includegraphics[width=0.5in]{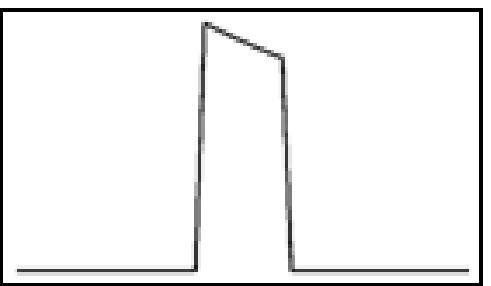} 
                         \\ \hline
                         
                                5 &     \includegraphics[width=0.5in]{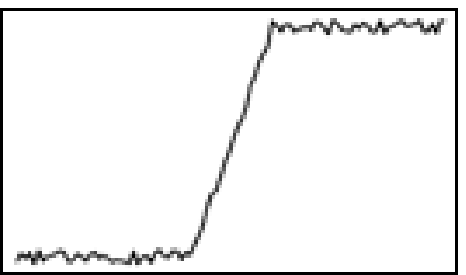} & 
                                 \includegraphics[width=0.5in]{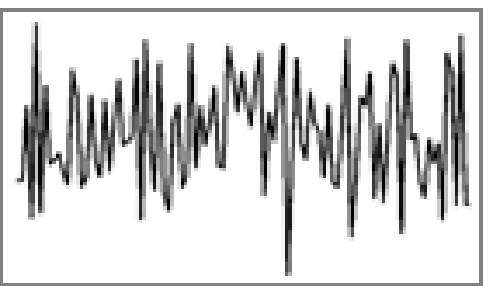}
                                 & 
                                 \includegraphics[width=0.5in]{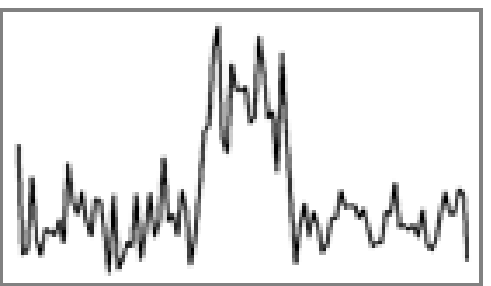}   \\ \hline
                                 
                                10 &    \includegraphics[width=0.5in]{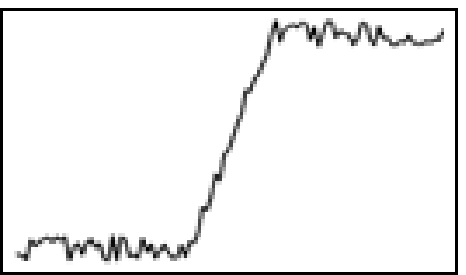} & 
                                  \includegraphics[width=0.5in]{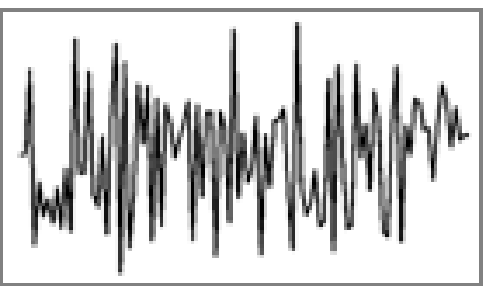}
                                  & 
                                  \includegraphics[width=0.5in]{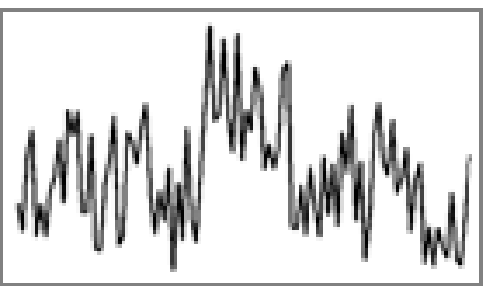}  \\ \hline
                                  
                                   20 & \includegraphics[width=0.5in]{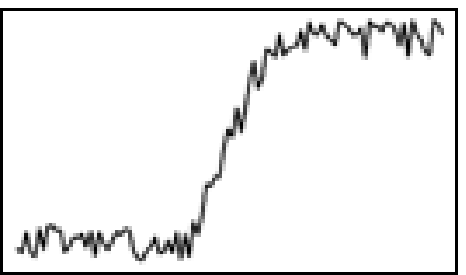} & 
                                   \includegraphics[width=0.5in]{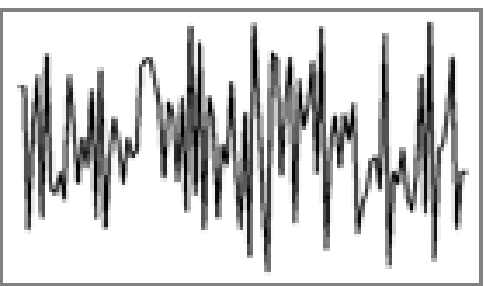}
                                   & 
                                   \includegraphics[width=0.5in]{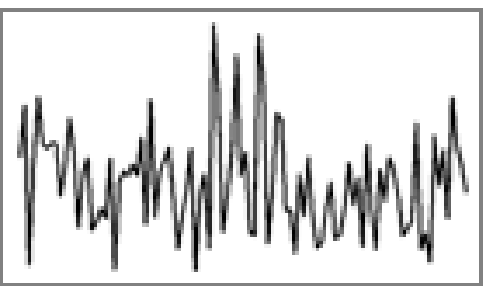}   \\ \hline
                         
                \end{tabular}
                \label{table3}
        \end{center}
\end{table}

Face images are used to study the quality of the edges detected by our proposed method in a realistic imaging condition. Performance evaluation of the proposed method is quantified using percentage correctness\footnote{The objective of the experiments are intended to objectively look at the quality of the edge features rather than with studying or comparing with existing face recognition methods.}, with the test data sets selected from face databases namely, ORL\footnote{\textbf{ORL} face images are acquired at different times with varying the lighting, facial expressions (open / closed eyes, smiling / not smiling) and facial details (glasses / no glasses). The background used is a dark homogeneous background with the subjects in an upright, frontal position (with tolerance for some side movement). }, AR\footnote{\textbf{AR Face Database} represents different image conditions such as facial expressions, illumination conditions, and occlusions (sun glasses and scarf)}, GeorgiaTech (GTech)\footnote{\textbf{Georgia Tech (GTech)} face images show frontal and/or tilted faces with different facial expressions, lighting conditions and scale. }, and Japanese Female Facial Expression (JAFFE)\footnote{\textbf{JAFFE} database contains 213 images of 7 facial expressions (6 basic facial expressions + 1 neutral) posed by 10 Japanese female models. Each image has been rated on 6 emotion adjectives by 60 Japanese subjects.},  These databases cover a wide range of  imaging conditions. Unless otherwise specified, the proposed method is evaluated against other commonly used edge detection methods such as Prewitt, Kirsch, SIS and Sobel using a minimum distance nearest neighbour classifier. 

\begin{figure}[hbtp]
\center
\subfigure[]{
  \includegraphics[width=40mm]{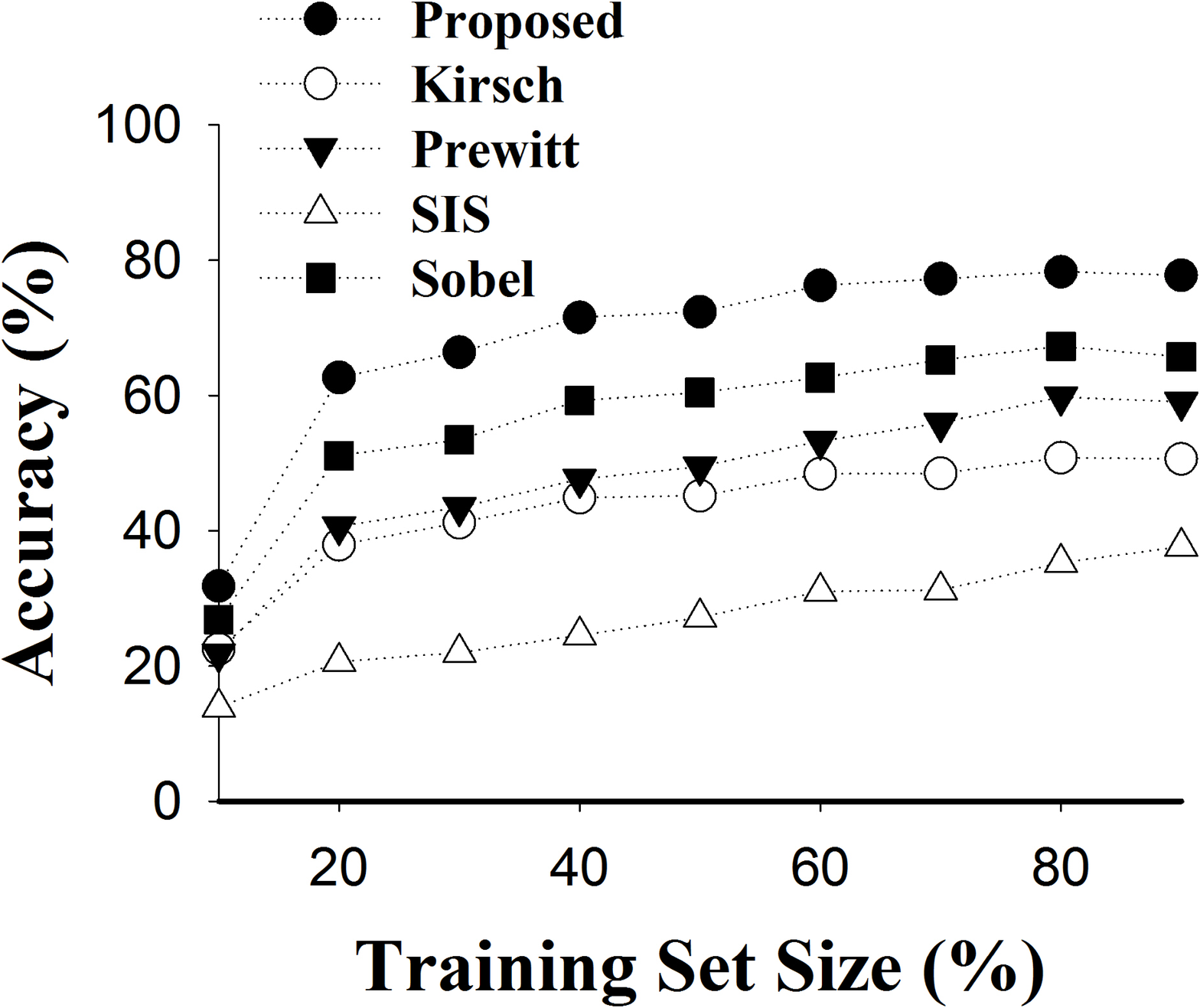}
   \label{fig3:subfig1}
   }\subfigure[]{
  \includegraphics[width=40mm]{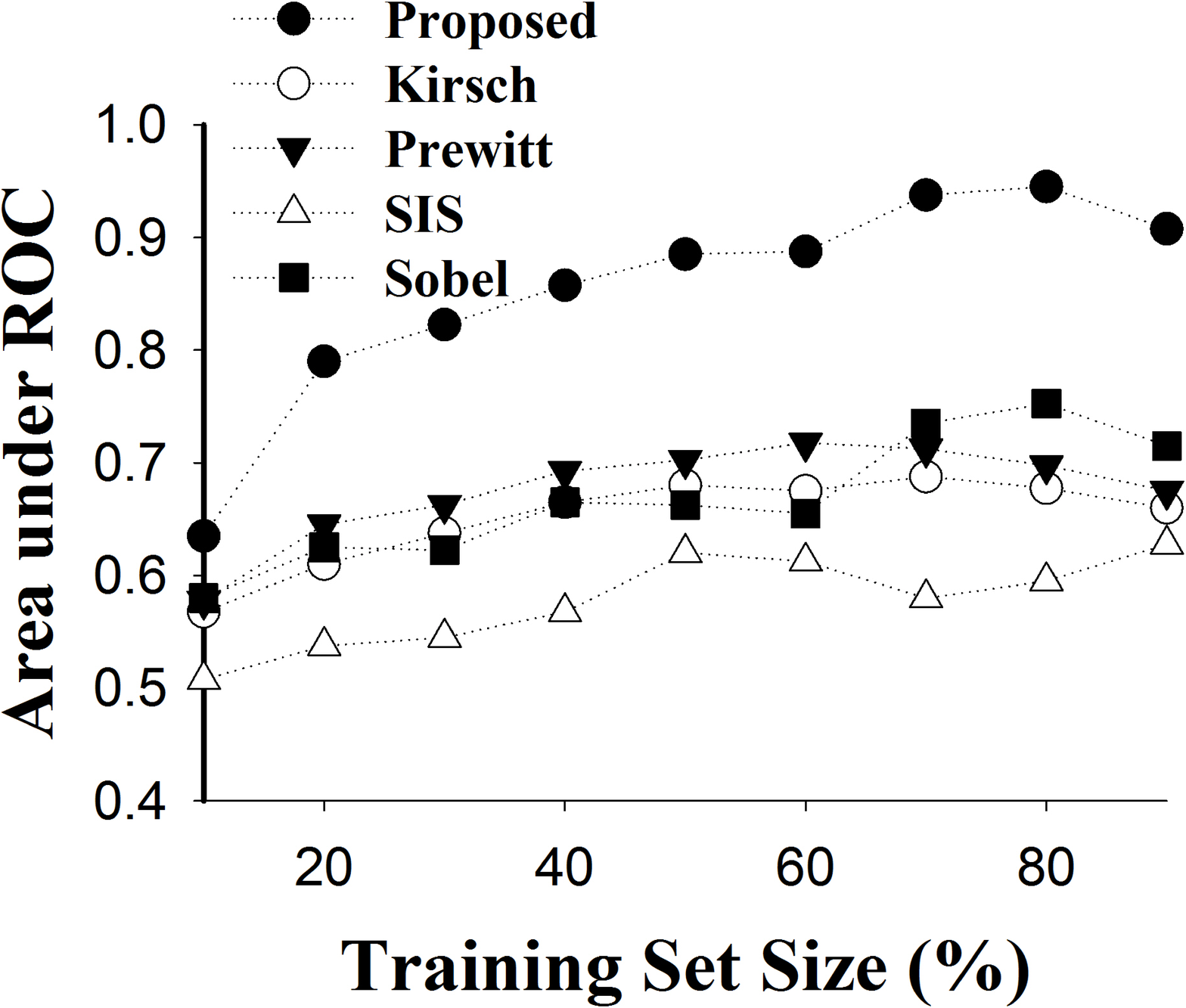}
   \label{fig3:subfig2}
   }
\caption{Average variation in accuracy (\%) (a), and area under ROC (b) for varying sizes of training set (\%) computed for ORL, AR, Georgia Tech, and JAFFE face databases calculated using a lazy classifier.}
\label{fig3}
\end{figure}

The recognition accuracy and area under receiver operator characteristics curve,   for varying size of the training set is illustrated with the help of Fig. \ref{fig3} using average performances from databases ORL, AR, GTech, and JAFFE. In a general, the proposed method shows a consistent improvement in robustness over the other methods. 

The comparisons of average recognition accuracy  on different classifiers are tabulated for proposed method benchmark  against Kirsch, Prewitt, SIS, and Sobel edge detection methods in Tables IV using databases ORL, AR, GTech, and JAFFE. The classification experiments are conducted by dividing the database into equal sized sets of training and test images.
The classifiers used in the recognition experiments  are IB-k\cite{10}, NNge\cite{9}, random forest\cite{11},  and SMO\cite{12}.
The results indicate statistically significant improvement in robustness of the proposed edge features over the existing edge models.  
\begin{table}[h]
\caption{Average recognition accuracy (\%)  for different classifiers  using face databases ORL, AR, Georgia Tech, and JAFFE}
\begin{tabular}{|c|c|c|c|c|}
\hline
\multirow{2}{*}{\textbf{Method}} & \multicolumn{4}{c|}{\textbf{Database}}                                                 \\ \cline{2-5} 
                                 & \textbf{IBk} & \textbf{NNge} & \textbf{RandomForest} & \textbf{SMO} \\ \hline
\textbf{Proposed}                &\textbf{76.1$\pm$6.4}        &\textbf{67.0$\pm$9.2} &\textbf{41.4$\pm$9.3} &\textbf{80.9$\pm$6.1}                   \\ \hline
\textbf{Kirsch}                     &46.8$\pm$8.5      &41.6$\pm$9.7  &32.2$\pm$8.4  &62.02$\pm$8.9                   \\ \hline
\textbf{Prewitt}                  &54.3$\pm$8.2        &44.9$\pm$7.8  &34.4$\pm$7.2  &61.6$\pm$8.1               \\ \hline
\textbf{SIS}                        &31.2$\pm$8.2      &25.6$\pm$7.2  &20.1$\pm$8.9              &39.5$\pm$8.8                      \\ \hline
\textbf{Sobel}                    &63.6$\pm$7.5           &50.1$\pm$11.1        &40.7$\pm$10.5  &77.1$\pm$8.5                  \\ \hline
\end{tabular}
                \label{table5}
\end{table}

\begin{figure}[hbtp]
\center
 \includegraphics[width=50mm]{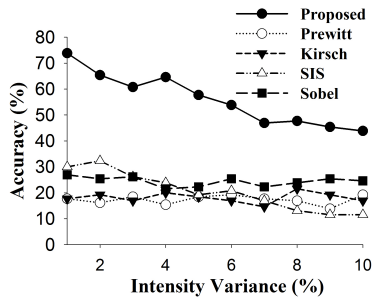}
 \caption{Average accuracy computed from example AR face database illustrating the impact of percentage intensity variability in pixels. Here,  x axis represents the maximum additive noise variance per pixel.}
\label{figure4}
\end{figure}

The proposed edge model contribute to reducing intra-regional intensity variance that in turn reflects improvements in  the recognition accuracy. For illustrating the effect, face databases are added with additive noise variance in increasing order. Figure 6 show significantly better tolerance towards intensity variance in comparison to other methods. The proposed theory preserves the edge information using fundamental principles from the psychology of vision processing. While the noise itself is not substantially removed as seen from Table \ref{table1}, \ref{table2}, \ref{table3}, and Fig 5, 6, while it does indicate that edges are preserved and that it can be processed further using filtering schemes to retrieve edges close to its non-noisy state.

\section{Conclusion}

In this paper, we presented the  idea of spatial stimuli gradient sketch model
inspired from the Fechner and Weber law, and Sheperd similarity measure that enables  relationships between the  image intensity and  psychological measurement space. The use of these concepts results in edges that are robust to noise in pixels intensity along the edges of different nature and overcome the limitations of edges based on primal sketch models. The robustness of the method is illustrated with the help of face recognition experiments on some of the standard benchmark databases that cover the wide range of possible edges in the digital images. The use of proposed method showed a statistically significant improvement in the recognition accuracies over the benchmark edge detection methods. The comparison  shown in Fig 6 indicate that while almost every benchmark edge detection methods fail to work as useful features in  a face recognition problem, the proposed edges show a higher level of tolerance to pixel noise levels. In future, the proposed spatial stimuli gradient sketch model can be used to extract more useful features for a wide range of problems such as image registration, image recognition, and image segmentation, where the use of edges is proven to be useful and higher levels of abstraction shown to improve the performances.


%

\ifCLASSOPTIONcaptionsoff
  \newpage
\fi



%

%



\end{document}